\documentclass[twoside,11pt]{article}
\usepackage[preprint]{dmlr2e}
\usepackage{booktabs}
\usepackage{amsmath}
\usepackage{microtype}
\usepackage{url}

\makeatletter
\def\@maketitle{\vbox{\hsize\textwidth
 \linewidth\hsize \vskip \beforetitskip
 {\begin{center} \Large\bf \@title \par \end{center}} \vskip \aftertitskip
 {\def\and{\unskip\enspace{\rm and}\enspace}%
  \def\addr{\small\it}%
  \def\email{\hfill\small\sc}%
  \def\name{\normalsize\bf}%
  \def\AND{\@endauthor\rm\hss \vskip \interauthorskip \@startauthor}
  \@startauthor \@author\@endauthor}
  \vskip \aftermaketitskip
}}
\makeatother
\ShortHeadings{CleanScore: Black-Box Benchmark Audits}{Opoku and Banahene}
\firstpageno{1}

\begin{document}

\title{CleanScore: Black-Box Benchmark Audits with Negative Controls and Sensitivity Bounds}

\author{\name Jeffery Opoku (corresponding author) \email jeffery.opoku01@utrgv.edu \\
       \addr School of Mathematical and Statistical Sciences\\
       The University of Texas Rio Grande Valley\\
       Edinburg, Texas, USA
       \AND
       \name David Banahene \\
       \addr Department of Biostatistics, Robert Stempel College of Public Health and Social Work\\
       Florida International University\\
       Miami, Florida, USA}

\maketitle
\begin{abstract}
Public benchmark scores reflect skill, prior exposure to the questions, or both, and for most models the training data are unknown. CleanScore is a black-box audit: the estimator needs only a correct-or-incorrect score for each form, not training data or weights. Each benchmark question becomes a parent item with one public form and two independently written fresh forms that keep its numbers, facts and answer. The audit reports an interval for the public-form advantage, not a verdict, and a private control bank with an explicit transport radius provides a proxy for the ordinary form gap that a rewrite alone can produce. A registered controlled-exposure experiment, a preliminary validation, detects planted exposure in one of two model families. A registered audit of five open models on 200 GSM8K and 200 ARC-Challenge items (ARC with stem-only rewrites, scored by option log probabilities) finds no exposure-consistent advantage, and two reruns declared before they were run (GSM8K at a longer token budget, ARC scored from generated letters) leave that conclusion unchanged. For four of the ten model-benchmark pairs the upper end of the interval is about five points or less, and for none is it above 21 points. The registered claim rule is much less sensitive than the primary interval: it flags a planted ten-point advantage in 5\% to 11\% of audits. Registered positive controls show what a null can mean. In four models, each trained under four seeds, items leaked into fine-tuning gained 25 points on average over unleaked items on the leaked wording and 19 points on paraphrases the model never saw, so a substantial share of the effect, estimated at 76\% (95\% interval 50\% to 113\%), is invisible to a paraphrase audit. A null on surface form says much less about contamination than the name suggests.
\end{abstract}


\section{Introduction}
Public benchmark scores shape model rankings, product claims, and research decisions. Their meaning is less clear when benchmark questions may have appeared in training data. For most closed models nobody outside the lab can inspect the training corpus, so a score can mix two things: the model's ability to solve the task, and an advantage from having seen the same or closely related material before.

Several tools address this. Clean-Eval rewrites evaluation sets, PaCoST compares model confidence on original and counterpart questions, ConStat corrects for difficulty with reference models, DICE inspects internal states, and DyePack plants markers before a benchmark is released. Most of them need something an ordinary API user does not have: token probabilities, model internals, reference models known to be clean, or control of the benchmark before it became public. Most also return a verdict, a flag or a p-value, without saying how large the inflation could be or how sure the auditor can be of that.

CleanScore asks a narrower question that a correct-or-incorrect score for each form can answer. For a fixed model and benchmark, is accuracy on the public form of a question higher than accuracy on independently written forms that need the same answer and the same reasoning? That difference can be observed, but it cannot be read directly: a fresh form can be slightly harder, easier or less natural than the original. The central problem is therefore to estimate the gap and to separate what the data identify from what depends on an assumption about form mismatch. CleanScore states that assumption, bounds it with a never-public control bank, and reports the conclusion across a visible range of transport radii. The estimator uses only those scores. Our GSM8K audit obtains them from generated text alone. Our ARC audit obtains them from option log probabilities, which some APIs do not return (Section 7.1); a second ARC readout that uses generated letters alone is in Section 7.3.

The paper makes five contributions.

\begin{enumerate}
\item A black-box matched-item estimand for the public-form advantage on a finite benchmark, with a conservative interval and a finite-population interval reported side by side, and a guardrail for small or sparse audits (Sections 3 and 6).
\item A sensitivity interval with an explicit transport radius from the control bank to the benchmark, and an exact placebo-label identity that checks the scoring pipeline (Section 4).
\item A registered controlled-exposure validation in two model families and a registered audit of five open models on GSM8K and ARC-Challenge, with every deviation reported (Sections 5 and 7).
\item A power calibration of the audit and of its registered claim rule, with an exact randomization test for small audits whose cost under form mismatch is measured, and the registered design of a larger audit that has not yet been run (Sections 7.6 and 7.7 and Appendix B).
\item Registered positive controls that measure how much of a real contamination effect a paraphrase audit can see, on GSM8K and on ARC (Section 8).
\end{enumerate}

Appendix B.9 (Table 27) lists every analysis in the paper by what was fixed before its results were seen.

The last result qualifies all the others, so we state it now. Four models from four families were each deliberately contaminated, under four training seeds, with the public wording of a hundred audited items. Leaking an item raised accuracy on the leaked wording by 25 points against uncontaminated control items, and raised accuracy on paraphrases the model had never seen by 19 points. A paraphrase audit sees only the difference, about 6 points. Figure 1 shows the split.

\begin{figure}[t]
\centering
\includegraphics[width=\textwidth]{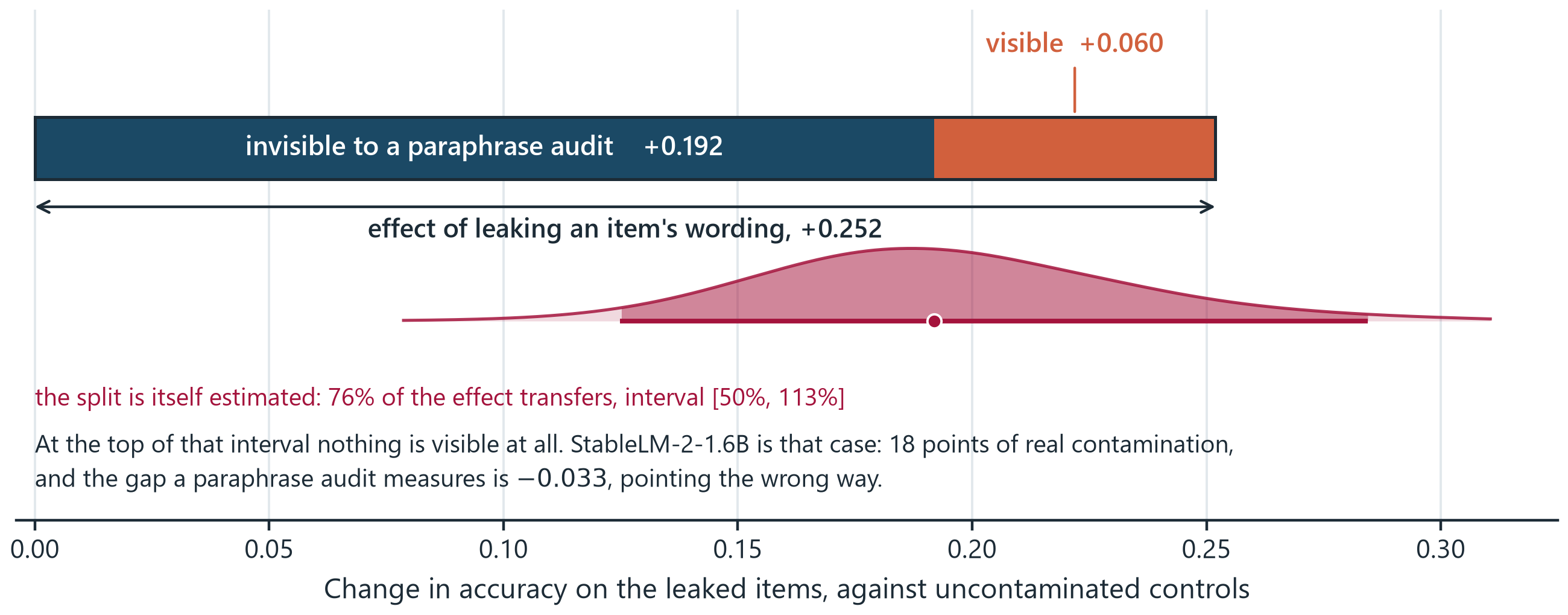}
\caption{The finding of Section 8.1, stated first because it limits what every other result can mean. Leaking an item's wording raises accuracy on that wording, and a substantial share of the gain carries over to paraphrases of the same item, where an audit that compares an original with a rewrite cannot see it. The estimated invisible share is 76\%, with an interval from 50\% to 113\%; at the top of that interval nothing is visible at all.}
\label{fig:1}
\end{figure}

The consequence reaches beyond CleanScore. Any audit that compares an original question with a rewrite of it measures only the visible part. A null from such an audit bounds surface-form inflation and bounds total contamination much more weakly than the label ``contamination audit'' implies, and we read the nulls of Section 7 that way. We measure the invisible share on GSM8K. On ARC a separate registered control shows the same blindness for a different reason: the rewrites of that benchmark keep the options fixed, so a memorized answer string survives them intact. Section 9 says what each result leaves open.

CleanScore does not claim to be the first paraphrase audit, the first contamination detector or the first to calibrate a paraphrase contrast against models with a known dose, and it does not show that a proprietary model used a particular dataset. It offers a reusable statement of uncertainty for the case where only a score for each form is available.

\section{Related work}
Clean-Eval builds filtered, paraphrased evaluation sets and shows why public benchmark forms can give optimistic results (Zhu et al., 2024). PaCoST compares model confidence on original and counterpart questions (Zhang et al., 2024). ConStat estimates benchmark-specific inflation by comparing a target model with reference models and reference benchmarks (Dekoninck et al., 2024a). DICE uses internal model states to detect in-distribution contamination during fine-tuning (Tu et al., 2024). DyePack places randomized markers in a benchmark before exposure and can give stronger evidence when that design is available (Cheng et al., 2025). Closed benchmarks such as MMLU-CF reduce future exposure risk but do not audit the public benchmarks already in use (Zhao et al., 2024).

Xiao and Cheng (2026) is the closest work in method. They reanalyse the item outputs released with ConStat, for 47 public models and for 74 Llama-2-7B models fine-tuned under known contamination conditions, on ARC, GSM8K, HellaSwag and MMLU. Their statistic is a model's accuracy on the original items minus its accuracy on their paraphrases, referenced to a clean base model. It rises with the contamination dose (+0.187 for leakage of the test set) and stays near zero for models trained only on the legitimate training split (-0.012). Their question is whether leaderboards are reordered. On HellaSwag, clean base models show a raw gap near 0.21 from the paraphrase alone, which is why a control is needed.

Neither the within-item contrast nor calibration against a known dose is new here. Four things differ. We run new inference on forms written for the audit, under a registration. The target is the public-form advantage on a finite benchmark, with an interval, not a ranking. A private control bank takes the place of a clean base model. And we leak the original wording and measure how much of the effect reaches paraphrases the model never saw, a quantity their paper does not report. Their condition that leaks the paraphrased test runs the other way, and its corrected gap is -0.091.

Sun et al. (2025) test twenty strategies for updating a benchmark on five evaluation sets and ten models. They report that no strategy that preserves meaning gives a significant gain in contamination resistance over the unmodified benchmark across all benchmarks. That agrees in direction with Section 8, where a substantial share of a leak's effect survives rewording. Fu et al. (2025) survey the detection literature, group the assumptions behind detectors into eight categories and test three of them. Membership inference methods built on those assumptions can perform like random guessing on data used in pretraining.

The closest work in intent attaches a formal guarantee to a contamination decision. Zhang et al. (2026) combine several detectors under an adaptive weighting scheme and control the false discovery rate among the samples a filter keeps, so a user can set in advance how much contamination to tolerate in an evaluation set. Their guarantee concerns which items survive a filter; ours concerns how large an advantage could be on the items that were kept. The two complement each other. Their method also needs per-sample detector scores, which an API user auditing a closed model does not have.

Oren et al. (2024) give a black-box test that needs no reference model and no control of the benchmark. A contaminated model finds a benchmark's canonical example order more likely than a shuffled one, and this yields a test with a provable guarantee from scored outputs alone. It is close to our exact test (Appendix B.2), and both rest on exchangeability, but of different kinds. Theirs is over the order of examples, a property of how the benchmark was assembled and plausible by construction. Ours is over which of a parent's forms is the original, and it fails whenever a rewrite is slightly harder than what it replaces. Their test shows that a benchmark was trained on. Ours, when it rejects, cannot separate contamination from form mismatch, so we use it only to strengthen a null. An audit that wants a positive finding from a black box should use theirs.

Yang et al. (2023) show that rephrasing or translating test items slips past n-gram decontamination, and that a 13B model trained on such variants matches far larger models. Our result is the mirror image: once such a model exists, an audit that compares an original with a rewrite cannot see a substantial share of the contamination either. Their concern is the decontamination pipeline and ours is the audit that follows it. Dekoninck et al. (2024b) show that several published detectors can be evaded cheaply, by contaminating a model in ways that leave the detector's statistic untouched. We read our nulls in the same spirit. CleanScore measures a surface-form advantage and reports how large it could be. It does not test whether a model is contaminated, and Section 8 measures how much of a planted contamination it fails to register.

Another line of work rebuilds a benchmark instead of rewording it. GSM1k writes new grade-school arithmetic problems matched to GSM8K in style and difficulty, and reports accuracy drops of up to 8\% for some model families and near-zero gaps for others (Zhang, Da, et al., 2024). That design measures dependence on the benchmark's whole distribution: its problem types, phrasing and answer magnitudes. CleanScore measures something narrower, the advantage attached to the exact wording of the published item. A model can be unaffected by rewording and still be helped by training on many problems of the same shape, so we report our nulls as statements about surface form and not as evidence that the models are clean.

The design borrows from two other literatures. Negative controls are an established way to reveal bias that a primary comparison cannot separate (Lipsitch et al., 2010), and the finite-population bounds are standard (Hoeffding, 1963; Serfling, 1974; Cochran, 1977). We do not claim these tools as new. The contribution is the audit design that connects them to matched forms, private controls and an explicit sensitivity analysis.

\section{Audit setup}
Figure 2 shows the whole procedure: where the forms come from, what the model is asked, and how the two estimands combine into the rule that decides whether a contamination claim may be made. This section defines the estimand and its intervals. Section 4 defines the control bank and the transport radius, and Section 7.1 records the settings used in the audit reported here.

\begin{figure}[t]
\centering
\includegraphics[width=\textwidth]{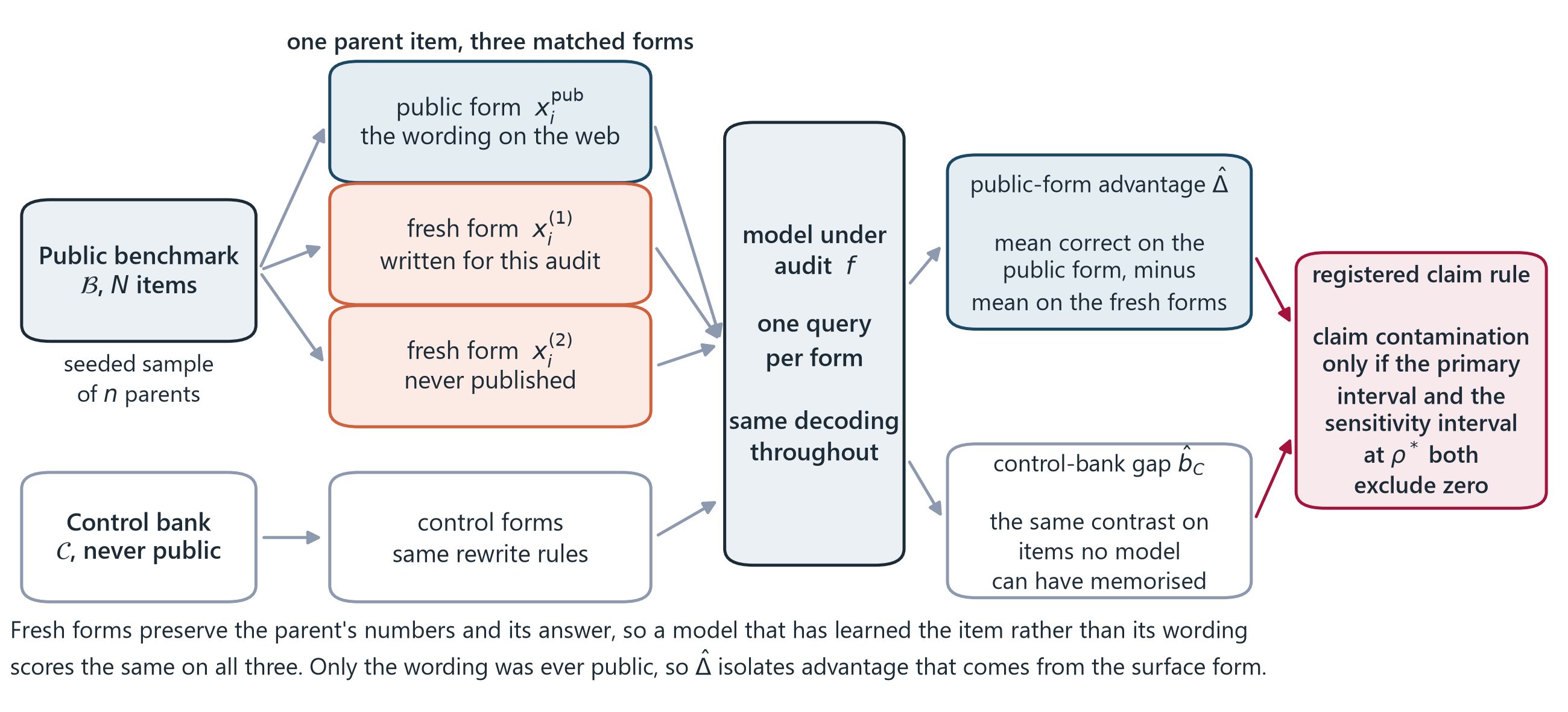}
\caption{The matched-form audit. Each sampled parent item is put to the model in its public form and in two fresh forms that keep the same numbers, facts, and answer; the public-form advantage is the difference in mean correctness between them. A never-public control bank of items no model can have memorized is scored the same way and supplies the form-mismatch term. The registered claim rule requires both intervals to exclude zero, so a form mismatch that the control bank reproduces, up to the transport radius, cannot on its own produce a contamination claim.}
\label{fig:2}
\end{figure}

\subsection{Parent items and matched forms}
Benchmark parent item $i$ has one public original form and $K$ fresh forms. All forms must have the same answer criterion and require the same reasoning. Let $Y_{iO}$ be correctness on the original and $Y_{iFk}$ correctness on fresh form $k$. Define

\begin{equation*}
D_i=Y_{iO}-\frac{1}{K}\sum_{k=1}^K Y_{iFk}.
\end{equation*}

The value $D_i$ lies between -1 and 1, and a positive value means the model did better on the public form of that parent.

Fresh forms stay private until all queries and scoring are done. The protocol calls for two checks before any model is queried. The first is mechanical and automated: answer preservation, preservation of every quantity the item depends on, and, for multiple-choice items, untouched options and key. The second is a reading by the authors for missing information, added or removed reasoning steps, and unnatural phrasing. Section 7.1 states which of the two had been completed, and when, for the forms used in this paper.

\subsection{Finite-benchmark target}
For a benchmark of $N$ parent items the target is

\begin{equation*}
\Delta_N=\frac{1}{N}\sum_{i=1}^N D_i.
\end{equation*}

If every item is audited, this mean has no item-sampling error. If only $n$ parents are sampled, the estimator is the sample mean, or its stratified weighted version.

\subsection{Dual uncertainty reporting}
Because $D_i \in [-1,1]$, a simple random sample without replacement gives the conservative half-width

\begin{equation*}
h_H=\sqrt{\frac{2\log(2/\alpha)}{n}}.
\end{equation*}

For a stratified audit with benchmark weights $W_h$ and sample sizes $n_h$ the conservative half-width becomes, by Proposition 2 in Appendix A,

\begin{equation*}
h_{H,\mathrm{str}}=
\sqrt{2\log(2/\alpha)\sum_h\frac{W_h^2}{n_h}}.
\end{equation*}

These bounds hold at any sample size and need no distributional assumption, but they can be too wide to be useful. We therefore also report the standard finite-population variance estimate

\begin{equation*}
\widehat V=
\sum_h W_h^2(1-f_h)\frac{s_h^2}{n_h},
\qquad f_h=\frac{n_h}{N_h},
\end{equation*}

and the large-sample interval

\begin{equation*}
\widehat\Delta_{\mathrm{str}}
\pm z_{1-\alpha/2}\sqrt{\widehat V}.
\end{equation*}

The two intervals answer the same sampling question and carry different guarantees. Both are always reported.

\section{Negative-control sensitivity analysis}
\subsection{Why the public-form gap is not enough}
Let $b_B$ be the form gap the same model would show on benchmark $B$ with no prior exposure, and let $\tau_B$ be the exposure-related part. Then

\begin{equation*}
\Delta_B=\tau_B+b_B.
\end{equation*}

The audit estimates $\Delta_B$. It never observes the counterfactual $b_B$, and human ratings of difficulty cannot bound a model-specific accuracy difference.

\subsection{Private control bank}
The control bank is never public before evaluation. Each control parent has one public-like first form and later fresh forms made under the same protocol as the target audit. Let $b_C$ be its average form gap. We use the transport assumption

\begin{equation*}
|b_B-b_C|\leq\rho.
\end{equation*}

The radius $\rho$ is not fitted to reach a desired conclusion. Results are shown over a grid fixed before the audit was unblinded.

If $I_\Delta = [L_\Delta, U_\Delta]$ covers $\Delta_B$ and $I_C = [L_C, U_C]$ covers $b_C$, the sensitivity interval is

\begin{equation*}
[L_\Delta-U_C-\rho,\;U_\Delta-L_C+\rho].
\end{equation*}

When the component intervals have error probabilities $\alpha_\Delta$ and $\alpha_C$, this interval covers $\tau_B$ with probability at least $1-\alpha_\Delta-\alpha_C$ under the transport assumption, provided both components have their stated coverage (Proposition 3, Appendix A). That proviso matters in practice. The control bank is composed, not sampled from a frame, so the interval on it is a heuristic, and Section 7.4 shows how much the conclusions depend on it.

When the audit parents and the control parents are independent samples, a single interval on the difference between the two gaps, with the two variances added, covers $\tau$ with the same probability and is about 38\% narrower at the 0.05 level and 35\% narrower at the 0.01 level the extension uses (Proposition 3b, Appendix A). It rests on a normal approximation for both components, so it cannot be used where the guardrail of Section 6.3 applies to either one. Every control bank of the first audit is small enough to fall under the guardrail, so that audit uses Proposition 3 throughout. The extension of Appendix B.8 has control banks large enough for Proposition 3b and falls back to Proposition 3 for a model whose control gaps are too sparse.

The structure is not special to auditing. When a method is calibrated on one population and applied to another, the guarantee holds up to an explicit mismatch term, and the useful question is how large that term must be before the conclusion changes. Conformal risk control under prompt drift and under tool-use drift is stated the same way, with a drift term added to the target risk (Opoku and Banahene, 2026a, 2026b). An audit cannot reweight its way out of the mismatch, so it reports the whole sensitivity curve.

\subsection{Placebo-label check}
For a private parent with (K+1) forms, one form is given the pseudo-original role uniformly at random after the responses are collected. If $J$ is that label,

\begin{equation*}
D^{\mathrm{pl}}(J)=Y_J-\frac{1}{K}\sum_{k\neq J}Y_k.
\end{equation*}

Averaging over all labels gives exactly zero. This checks the code, the scoring and the form-role handling. We confirmed the identity on 500 random four-form outcome vectors, with a largest residual below $6 \times 10^{-17}$. It does not show that a historically public original is exchangeable with a new form.

\section{Controlled-exposure validation}
The validation trains a small open model under four mixtures and asks whether the audit sees what was planted. The four conditions are a clean control with unrelated instruction data; a related-task control with arithmetic practice that shares no item with the audit bank; exact exposure, which adds the public forms of a random half of the 120 audit parents to the related-task mixture; and paraphrase exposure, which adds fresh forms of the same half instead. Every condition uses 1,200 training records and three epochs, and each is trained under ten seeds, giving 40 model fits per family.

The analysis was locked before any condition was run (\texttt{PAPER10\_\allowbreak ANALYSIS\_\allowbreak LOCK.md}). The unit of analysis is the trained model run, not the question row. For each seed and condition the parent gaps are averaged into one public-form advantage, and the two registered contrasts are the seed-paired differences of exact exposure and of paraphrase exposure against the related-task control. The lock also stated that exposure to one arithmetic item may help related unexposed items, so exposed-versus-unexposed comparisons inside an arm are diagnostics and not causal estimates. The registered model was Qwen2.5-0.5B-Instruct. An independent replication in a second family, SmolLM2-360M-Instruct, used the same design, seeds and lock. All 80 fits completed on Kaggle T4 hardware, and the run artifacts are archived.

\begin{table}[t]
\centering
\small
\caption{Registered seed-paired contrasts in the controlled-exposure validation (ten seeds per family).}
\label{tab:1}
\resizebox{\ifdim\width>\textwidth \textwidth\else \width\fi}{!}{%
\begin{tabular}{llrlrr}
\toprule
\textbf{Family} & \textbf{Contrast} & \textbf{Mean} & \textbf{95\% CI} & \textbf{Seeds positive} & \textbf{p (two-sided)} \\
\midrule
SmolLM2-360M & Exact exposure minus related-task & +0.020 & [+0.006, +0.034] & 9/10 & 0.010 \\
SmolLM2-360M & Paraphrase exposure minus related-task & -0.002 & [-0.009, +0.005] & 3/10 & 0.61 \\
Qwen2.5-0.5B & Exact exposure minus related-task & +0.043 & [-0.018, +0.104] & 7/10 & 0.14 \\
Qwen2.5-0.5B & Paraphrase exposure minus related-task & -0.015 & [-0.045, +0.016] & 3/10 & 0.31 \\
\bottomrule
\end{tabular}}
\end{table}

Table 1 gives the contrasts. In both families the exact-exposure contrast is positive and the paraphrase-exposure contrast is not. In SmolLM2-360M the exact-exposure contrast is clear: nine of ten seeds are positive and the interval excludes zero. In the registered model, Qwen2.5-0.5B, the contrast has the same direction and a larger point estimate, but its interval includes zero because the model's run-to-run variation is about four times larger (clean-arm standard deviation across seeds 0.061 against 0.016). We report it as direction-consistent and underpowered. Detection of planted exposure therefore rests on one of the two families.

Inside the exact-exposure arm, the exposed-versus-unexposed difference is +0.018 in Qwen2.5-0.5B and -0.005 in SmolLM2-360M. The advantage spreads across the whole item bank instead of concentrating on the 60 planted parents. The lock named this spillover in advance: exposure to public arithmetic items teaches the public surface form of the whole bank. It also means that an audit that samples unexposed items from a contaminated benchmark can still see part of the effect. The decision rule in the lock asked for a measurable exact-exposure signal and for well-behaved controls before the design was applied to a real benchmark, and both conditions were met.

\section{Design simulation}
\subsection{Sampling coverage}
We generated finite populations of 2,400 bounded parent-level gaps and sampled them repeatedly without replacement, comparing the conservative Hoeffding interval with the finite-population Wald interval over 5,000 replications for simple and stratified sampling (Table 2). The conservative interval covered every replication, but its width makes it weak for moderate gaps. The Wald interval stayed close to the nominal 95\% and was much narrower. This is a design result, not evidence about contamination in a real model.

\begin{table}[t]
\centering
\small
\caption{Coverage and width of the two intervals in the finite-population calibration, 5,000 replications per design.}
\label{tab:2}
\resizebox{\ifdim\width>\textwidth \textwidth\else \width\fi}{!}{%
\begin{tabular}{lrrrrr}
\toprule
\textbf{Design} & \textbf{Sample size} & \textbf{Hoeffding coverage} & \textbf{Hoeffding half-width} & \textbf{Wald coverage} & \textbf{Mean Wald half-width} \\
\midrule
Simple random sample & 120 & 1.0000 & 0.2480 & 0.9476 & 0.0625 \\
Simple random sample & 400 & 1.0000 & 0.1358 & 0.9506 & 0.0321 \\
Simple random sample & 800 & 1.0000 & 0.0960 & 0.9522 & 0.0203 \\
Simple random sample & 1,200 & 1.0000 & 0.0784 & 0.9468 & 0.0144 \\
Stratified sample & 400 & 1.0000 & 0.1358 & 0.9488 & 0.0329 \\
\bottomrule
\end{tabular}}
\end{table}

\subsection{Transport-radius failure}
The sensitivity interval is valid only when the stated radius is large enough, and a separate 3,000-replication stress grid shows what happens when it is not. With a true control-to-benchmark mismatch of 0.05 and 800 sampled items in each bank, empirical coverage was 0.445 when the assumed radius was zero, 0.931 when it was 0.025, and 0.9997 when it matched the true value 0.05. The mean interval widths were about 0.095, 0.145 and 0.195. A larger radius protects coverage and gives a wider conclusion. That tradeoff cannot be removed by wording.

\subsection{Small and sparse audits: the guardrail}
We tested 36 simple-random-sampling settings built from six bounded outcome shapes (sparse, near-null, high-variance two-point, low-variance, balanced and right-skewed), two benchmark sizes, and audit fractions of 5\%, 20\% and 50\%, with 2,000 audits per setting. The conservative interval covered at least 0.9935 in every setting. Wald coverage ranged from 0.9145 to 0.9575, with six settings below 0.94 and two below 0.93. The worst case was a 50-item audit of a sparse 1,000-item population. Three further designs with strongly imbalanced strata gave Wald coverage of 0.9440 under proportional allocation, 0.9475 under equal allocation and 0.9490 when the rare stratum was oversampled.

We then checked the worst case without Monte Carlo error. For four discrete finite populations we enumerated every possible sample composition and weighted it by its exact multivariate hypergeometric probability. In the sparse 1,000-item population with a 50-item audit, exact Wald coverage was 0.9127, close to the simulated 0.9145. It rose to 0.9467 at 200 audited items and 0.9499 at 500. The conservative interval had at least 0.9968 exact coverage in every enumerated setting. The small-sparse warning is therefore not a simulation artifact.

This sets the operating rule. The Wald interval is a secondary precision summary. When fewer than 200 parents are audited, or fewer than 30 sampled parents have a nonzero gap, the distribution-free interval is the primary result. Both intervals stay visible in every analysis. The thresholds are practical guardrails supported by these tests and are not universal cutoffs. The registration fixed them before any audited model was queried.

\section{Real-benchmark audit}
\subsection{Registered protocol}
The audit protocol was locked before any audited model saw an audit item (\texttt{REAL\_\allowbreak AUDIT\_\allowbreak REGISTRATION.md}). We used two benchmarks. The primary one is the GSM8K test split (N = 1,319). The secondary, held-out domain is the ARC-Challenge test split restricted to four-option items (N = 1,165). From each frame we drew 200 parents without replacement with one seeded generator. For every parent, two fresh forms were written under fixed rules. For GSM8K, change the names, entities and sentence structure but keep every numeric quantity, the logical structure, the units and the gold answer. For ARC, rewrite the question stem only and keep the four options and the key verbatim. The ARC arm is therefore a stem audit: it cannot see memorization attached to the options (Section 8.2).

A language model from a provider and family different from every audited model drafted the fresh forms. A deterministic script then checked each one: the multiset of numeric tokens, the gold answer, length bounds and, for ARC, option integrity and a token-dissimilarity threshold against the original stem. Twelve ARC rewrites failed the dissimilarity check on the first pass and were rewritten before any model was queried. The final bank has 600 GSM8K forms and 600 ARC forms.

The two checks of Section 3.1 ran at different times. The mechanical check ran on every form before any model was queried, and it can be reproduced from the released bank. Both authors did the reading, and it was finished after the results were produced. It could not have changed which items were audited, only what is now known about them. It was also prioritized: we worked from a ranked packet released with the paper, which lists the parents most likely to be defective together with every GSM8K rewrite at or above the dissimilarity threshold, and we did not read all 1,500 forms. The reading found no form that changes its parent's answer, drops information the question needs, or adds a hint. It found one defect, an ungrammatical ARC rewrite, which is left in the frozen bank (Appendix B.3). We therefore call the forms mechanically checked and partly read, not verified.

We composed two private control banks from scratch, and neither has ever been public: 60 GSM8K-style word problems and 40 ARC-style science questions, each with a public-like first form and two fresh forms written under the same rules (300 forms). No audited model can have seen them, so they estimate each model's ordinary form gap $b_C$ under this writing protocol. Table 3 traces the items from the frames to the analysis.

\begin{table}[t]
\centering
\small
\caption{Item flow of the registered audit, per model.}
\label{tab:3}
\resizebox{\ifdim\width>\textwidth \textwidth\else \width\fi}{!}{%
\begin{tabular}{lrr}
\toprule
\textbf{Stage} & \textbf{GSM8K} & \textbf{ARC-Challenge} \\
\midrule
Frame: test split (four-option items for ARC) & 1,319 & 1,165 \\
Parents drawn without replacement & 200 & 200 \\
Fresh forms written, two per parent & 400 & 400 \\
Forms scored: original plus two fresh forms per parent & 600 & 600 \\
Parents in the analysis, none excluded & 200 & 200 \\
Private control parents, each with a public-like and two fresh forms & 60 & 40 \\
Control forms scored & 180 & 120 \\
Defects found by the authors' reading, left in the bank & 0 & 1 \\
\bottomrule
\end{tabular}}
\end{table}

We registered five open-weight models with pinned revisions: Qwen2.5-1.5B-Instruct, Qwen2.5-7B-Instruct, Phi-3-mini-4k-instruct, SmolLM2-1.7B-Instruct and OLMo-2-1124-7B-Instruct. The last is included because its pretraining and post-training mixtures are public, so one model in the panel has data provenance a reader can check independently of the audit. Each model answered each GSM8K item once, by greedy decoding through its chat template with a fixed instruction and a 320-token budget, and answers were scored by exact numeric match after a registered parser. ARC items were scored without generation, by the character-normalized log probability of each option text given the stem. The ARC arm therefore needs option log probabilities, which some APIs do not return, and is gray-box in practice. The GSM8K arm needs generated text only. Section 7.3 adds a second ARC readout that uses generated letters alone.

The registered inference for each model and benchmark is the sample mean of the parent gaps $D_i$; the Hoeffding interval; the finite-population Wald interval; the guardrail of Section 6.3 (with every audit at exactly 200 parents, only the nonzero-gap clause can bind); a Wald interval for $b_C$ from the control bank; and the Proposition 3 sensitivity interval for the exposure-related component over the grid $\rho$ in {0, 0.01, 0.02, 0.05, 0.10}, with $\rho^*$ = 0.02 as the headline radius. A pair is reported as showing an exposure-consistent public-form advantage only if the primary interval for $\Delta$ excludes zero and the sensitivity interval at $\rho^*$ excludes zero. The placebo-label identity is computed on every control parent.

\subsection{Primary results}
All five models completed under the registered settings. Table 4 gives the primary analysis, and Figure 3 plots the ten estimates with their control-bank gaps.

\begin{table}[t]
\centering
\small
\caption{Registered real-benchmark audit, all five models. $\hat{\Delta}$ is the public-form advantage over 200 sampled parents. The half-width is the primary interval: the finite-population Wald interval, or the Hoeffding bound (marked H) where the nonzero-gap guardrail applies. $b_C$ is the control-bank form gap with its Wald half-width. No model-benchmark pair meets the registered claim rule.}
\label{tab:4}
\resizebox{\ifdim\width>\textwidth \textwidth\else \width\fi}{!}{%
\begin{tabular}{llrrrrrl}
\toprule
\textbf{Model} & \textbf{Benchmark} & \textbf{Public score} & \textbf{Fresh score} & \textbf{$\hat{\Delta}$} & \textbf{Half-width} & \textbf{Nonzero gaps} & \textbf{$b_C$ (control)} \\
\midrule
OLMo-2-7B & GSM8K & 0.720 & 0.705 & +0.015 & 0.045 & 58 & +0.017 $\pm$ 0.081 \\
OLMo-2-7B & ARC & 0.625 & 0.608 & +0.018 & 0.040 & 40 & +0.013 $\pm$ 0.102 \\
SmolLM2-1.7B & GSM8K & 0.475 & 0.430 & +0.045 & 0.058 & 83 & 0.000 $\pm$ 0.119 \\
SmolLM2-1.7B & ARC & 0.485 & 0.470 & +0.015 & 0.192 (H) & 26 & -0.088 $\pm$ 0.110 \\
Phi-3-mini & GSM8K & 0.820 & 0.803 & +0.018 & 0.035 & 35 & +0.008 $\pm$ 0.049 \\
Phi-3-mini & ARC & 0.595 & 0.598 & -0.003 & 0.192 (H) & 27 & 0.000 $\pm$ 0.105 \\
Qwen2.5-1.5B & GSM8K & 0.475 & 0.500 & -0.025 & 0.055 & 80 & +0.033 $\pm$ 0.093 \\
Qwen2.5-1.5B & ARC & 0.495 & 0.503 & -0.008 & 0.029 & 32 & +0.025 $\pm$ 0.093 \\
Qwen2.5-7B & GSM8K & 0.710 & 0.715 & -0.005 & 0.047 & 59 & -0.017 $\pm$ 0.052 \\
Qwen2.5-7B & ARC & 0.590 & 0.558 & +0.033 & 0.036 & 38 & +0.025 $\pm$ 0.086 \\
\bottomrule
\end{tabular}}
\end{table}

Under the registered rule, no model shows an exposure-consistent public-form advantage on either benchmark. The ten point estimates are small and of both signs. The largest, Qwen2.5-7B on ARC at +0.033, has a half-width of 0.036. The placebo-label identity returned exactly zero for every model and domain, so the scoring pipeline and the pseudo-original role assignment add no asymmetry.

How tight is each null? The upper end of the primary interval (the estimate plus its half-width) is about 0.05 or less for four pairs: Qwen2.5-1.5B on ARC and on GSM8K, Qwen2.5-7B on GSM8K, and Phi-3-mini on GSM8K. For Phi-3-mini on GSM8K, +0.018 with half-width 0.035, the upper end of the interval is about five points. Four pairs lie between 0.057 and 0.103. The two pairs where the nonzero-gap guardrail applies, SmolLM2-1.7B and Phi-3-mini on ARC, have upper ends of 0.207 and 0.190. OLMo-2-7B, the one model whose training data are public, gives +0.015 and +0.018, in the same range as the models whose data are not documented. On ARC the rewrites rarely change a log-probability scorer's choice: only 26 to 40 of 200 parents show any gap. The sensitivity intervals for the exposure-related component $\tau$ at $\rho^*$ = 0.02 include zero for every pair, and their widths, from 0.23 to 0.44, are set by the control-bank intervals (Section 7.4).

\begin{figure}[t]
\centering
\includegraphics[width=\textwidth]{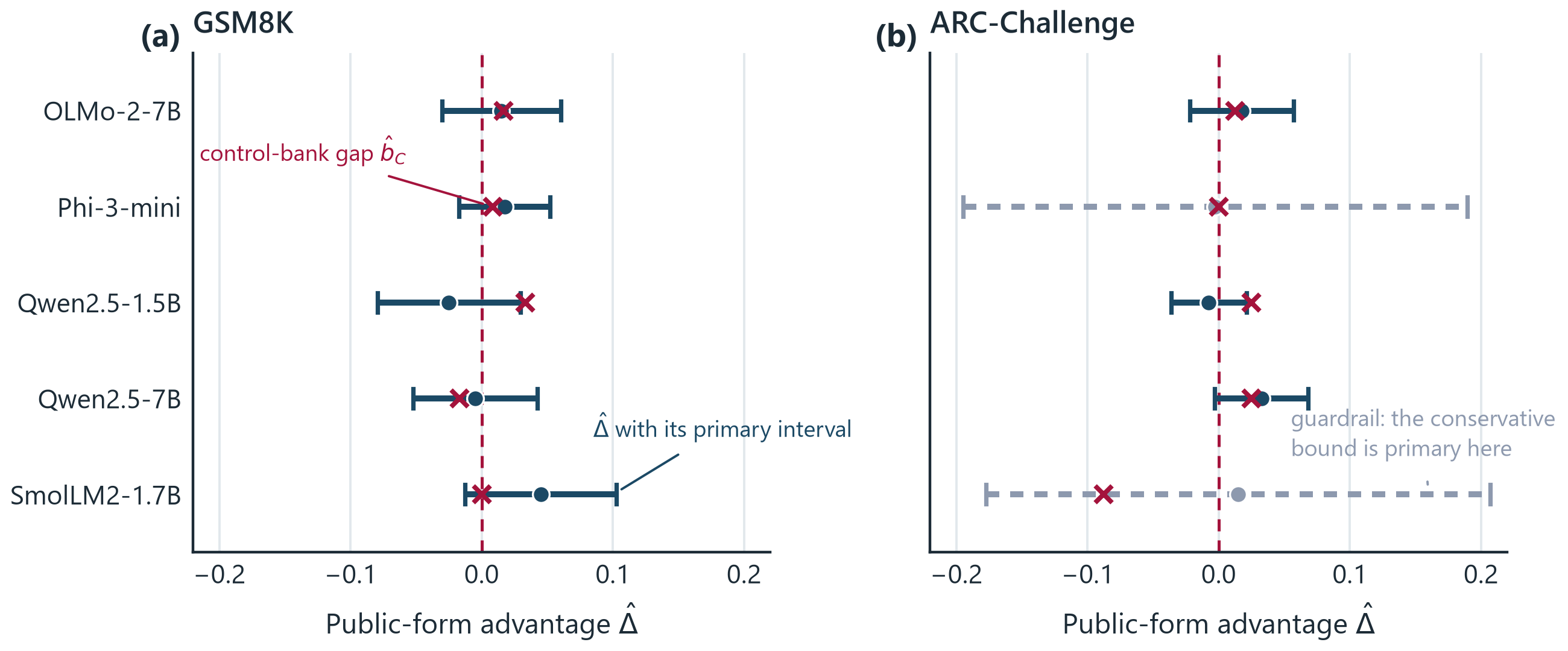}
\caption{Public-form advantage with its primary interval for every model-benchmark pair, with the control-bank gap $b_C$ marked separately. Dotted intervals are the two pairs where the registered nonzero-gap guardrail makes the conservative bound primary. Every interval covers zero.}
\label{fig:3}
\end{figure}

\subsection{Answer length, answer format and the ARC readout}
The saved outputs showed a problem with the registered generation budget. The two Qwen models hit the 320-token cap on 40.7\% (1.5B) and 29.8\% (7B) of GSM8K forms, and they wrote the registered answer phrase on only 2.7\% and 8.8\% of forms, using a boxed LaTeX format instead. OLMo-2-7B truncated on 25.3\%, SmolLM2-1.7B on 14.5\% and Phi-3-mini on 11.3\%. Truncation censors the outcome on the items that need long reasoning, and a rewrite could change how long a model writes. A check run after the results were known, and not registered, compares the two kinds of form for the three models that were not rerun (OLMo-2-7B, SmolLM2-1.7B, Phi-3-mini). Truncation on the public form and on the fresh forms is 24.5\% and 25.8\%, 13.0\% and 15.3\%, and 10.5\% and 11.8\%. Keeping only the parents with no truncated form leaves 120, 139 and 161 parents and moves the audit estimate from +0.015 to -0.008, from +0.045 to +0.043 and from +0.018 to +0.016. If every truncated outcome were allowed to be right or wrong in the worst way for the estimate, the bounds would be [-0.245, +0.258], [-0.105, +0.178] and [-0.098, +0.125], too wide to settle anything. Only a rerun settles it, and Amendment 2 below is that rerun. The truncation also explains why the Qwen public scores in Table 4 sit well below published GSM8K figures.

Before any rerun, Amendment 1 was logged: rerun the two Qwen models with a 768-token budget and a parser that also accepts a boxed answer, and report the result beside the registered primary, never instead of it. Table 5 gives the outcome. Truncation falls to zero, Qwen2.5-7B's score rises from 0.710 to 0.930 (close to its published figure) and Qwen2.5-1.5B's from 0.475 to 0.695. Both gaps stay small and negative and neither interval excludes zero, so the null is not an artifact of truncation. The ARC results in the rerun are identical to the primary run form by form, as expected, since ARC scoring involves neither generation nor the parser. The other three models were rerun later, under Amendment 2. An audit protocol should let the model finish its answer and should check format compliance before scoring.

\begin{table}[t]
\centering
\small
\caption{Amendment 1: the two Qwen models on GSM8K with a 768-token budget and a boxed-answer parser. Truncation falls to zero. ARC is unaffected by both changes and was reproduced bit-for-bit.}
\label{tab:5}
\resizebox{\ifdim\width>\textwidth \textwidth\else \width\fi}{!}{%
\begin{tabular}{llrrrrr}
\toprule
\textbf{Model} & \textbf{Setting} & \textbf{Public score} & \textbf{Fresh score} & \textbf{$\hat{\Delta}$} & \textbf{Half-width} & \textbf{Truncated} \\
\midrule
Qwen2.5-7B & Registered (320 tokens) & 0.710 & 0.715 & -0.005 & 0.047 & 29.8\% \\
Qwen2.5-7B & Amendment 1 (768 tokens) & 0.930 & 0.938 & -0.008 & 0.192 (H) & 0.0\% \\
Qwen2.5-1.5B & Registered (320 tokens) & 0.475 & 0.500 & -0.025 & 0.055 & 40.7\% \\
Qwen2.5-1.5B & Amendment 1 (768 tokens) & 0.695 & 0.735 & -0.040 & 0.054 & 0.0\% \\
\bottomrule
\end{tabular}}
\end{table}

Amendment 2 addresses the two points that the registered run and Amendment 1 left open: three models that had not been rerun at a longer budget, and an ARC arm that needs option log probabilities. It was declared before any model was queried on either arm and changes no form, bank, interval or claim rule. Its results sit beside the registered primary without replacing it. Table 6 gives both arms.

Arm A reruns GSM8K for OLMo-2-7B, SmolLM2-1.7B and Phi-3-mini at 768 tokens with the Amendment 1 parser. Truncation falls to 0.2\%, 2.0\% and 0.0\% of audit forms. The amendment calls a rerun adequate when at most 1\% of audit forms are truncated, so SmolLM2-1.7B is only partial: 12 of its 600 forms still reach the budget, one public form and 11 fresh forms. The public scores of OLMo-2-7B and Phi-3-mini rise from 0.720 to 0.865 and from 0.820 to 0.875. The public score of SmolLM2-1.7B moves from 0.475 to 0.485. The estimates are +0.008, +0.038 and -0.003, against +0.015, +0.045 and +0.018 in the registered run. No primary interval excludes zero and no pair meets the claim rule. Phi-3-mini now has only 22 parents with a nonzero gap, so the guardrail makes the Hoeffding bound primary and its half-width rises from 0.035 to 0.192. Its Wald half-width is 0.032 and its empirical Bernstein half-width 0.155. The 320-token nulls for these three models therefore survive a budget that lets nearly every answer finish.

Arm B scores ARC from the letter that each model generates, with a fixed prompt, a fixed parser and no re-prompting. It needs generated text only. Every model is interpretable: no reply to an audit form was invalid. Public-form scores rise from 0.590 to 0.900 for Qwen2.5-7B and from 0.595 to 0.880 for Phi-3-mini. The five estimates are +0.010, -0.008, +0.003, +0.008 and -0.010, and no pair meets the claim rule. Four of the five pairs have fewer than 30 parents with a nonzero gap, so the guardrail makes the Hoeffding bound primary, with half-width 0.192. The arm therefore gives a second readout of the same forms but little added precision. Under this readout OLMo-2-7B and SmolLM2-1.7B both show a public-like advantage of +0.088 on the control bank, and the Wald interval for each excludes zero (half-widths 0.078 and 0.069). Under the registered log-probability readout, no ARC control interval excluded zero. A control gap therefore depends on the readout as well as on the forms, which supports treating the composed control banks as informal (Section 7.4). For all ten pairs of the two arms, the sensitivity interval for $\tau$ at $\rho^*$ still includes zero.

\begin{table}[t]
\centering
\small
\caption{Amendment 2, reported beside the registered primary. Arm A reruns GSM8K at 768 tokens (the two Qwen rows are the Amendment 1 runs). Arm B scores ARC from the letter the model generates. ``Registered'' is the $\hat{\Delta}$ of the registered primary for the same model and benchmark. The half-width is that of the primary interval, with (H) where the nonzero-gap guardrail makes the distribution-free bound primary. The next-to-last column is the share of audit forms truncated at the budget (Arm A) or with an invalid reply (Arm B).}
\label{tab:6}
\resizebox{\ifdim\width>\textwidth \textwidth\else \width\fi}{!}{%
\begin{tabular}{llrrrrrrl}
\toprule
\textbf{Model} & \textbf{Setting} & \textbf{Registered $\hat{\Delta}$} & \textbf{Public score} & \textbf{Fresh score} & \textbf{$\hat{\Delta}$} & \textbf{Half-width} & \textbf{Truncated or invalid} & \textbf{Status} \\
\midrule
OLMo-2-7B & GSM8K, 768 tokens & +0.015 & 0.865 & 0.858 & +0.008 & 0.035 & 0.2\% & adequate \\
SmolLM2-1.7B & GSM8K, 768 tokens & +0.045 & 0.485 & 0.448 & +0.038 & 0.060 & 2.0\% & partial \\
Phi-3-mini & GSM8K, 768 tokens & +0.018 & 0.875 & 0.878 & -0.003 & 0.192 (H) & 0.0\% & adequate \\
Qwen2.5-1.5B & GSM8K, 768 tokens & -0.025 & 0.695 & 0.735 & -0.040 & 0.054 & 0.0\% & adequate \\
Qwen2.5-7B & GSM8K, 768 tokens & -0.005 & 0.930 & 0.938 & -0.008 & 0.192 (H) & 0.0\% & adequate \\
OLMo-2-7B & ARC, generated letter & +0.018 & 0.775 & 0.765 & +0.010 & 0.032 & 0.0\% & interpretable \\
SmolLM2-1.7B & ARC, generated letter & +0.015 & 0.605 & 0.613 & -0.008 & 0.192 (H) & 0.0\% & interpretable \\
Phi-3-mini & ARC, generated letter & -0.003 & 0.880 & 0.878 & +0.003 & 0.192 (H) & 0.0\% & interpretable \\
Qwen2.5-1.5B & ARC, generated letter & -0.008 & 0.785 & 0.778 & +0.008 & 0.192 (H) & 0.0\% & interpretable \\
Qwen2.5-7B & ARC, generated letter & +0.033 & 0.900 & 0.910 & -0.010 & 0.192 (H) & 0.0\% & interpretable \\
\bottomrule
\end{tabular}}
\end{table}

\subsection{What the control bank buys}
In the registered run, nine of the ten control banks gave gaps within 0.033 of zero (Table 4, last column). The tenth, SmolLM2-1.7B on ARC, gave -0.088, a gap discussed below. For the nine, the sensitivity interval for $\tau$ is essentially the audit interval widened by the control interval and the radius. To pay for two coverage events, the registered analysis splits the error budget and builds both components at $\alpha_\Delta$ = $\alpha_C$ = 0.025, so the combined interval has 95\% coverage if both components have their stated coverage. The composed control bank cannot guarantee that for its own component, as discussed below. These components are wider than the $\alpha$ = 0.05 half-widths in Table 4, and substituting the table's numbers into Proposition 3 gives an interval that is too narrow.

The control intervals are wide, from 0.049 to 0.119, because the banks hold 40 to 60 parents against 200 in the audit. Even where the audit interval has half-width 0.035, the sensitivity interval spans more than 0.2. A control bank should be at least as large as the audit sample, and its size should be chosen from the audit's target precision. A bank of 200 parents would bring the control half-width down to the audit's and roughly halve the width of the $\tau$ interval.

Two features of the registered construction need stating. First, the $\Delta$ component of the sensitivity interval is the Wald interval for every pair, including the two ARC pairs where the guardrail makes the distribution-free bound primary. Those pairs are the sparse regime in which Section 6.3 found Wald under-covering, and the error makes excluding zero easier. Recomputing them with the distribution-free component at the same $\alpha$/2 roughly doubles the width, from 0.355 to 0.711 for SmolLM2-1.7B and from 0.345 to 0.700 for Phi-3-mini. Neither excludes zero either way. An audit whose claim rule could actually fire should use the guardrail bound in both places. Second, the control banks are composed, not drawn from a frame, and are small enough that the guardrail would apply to them as well. Table 7 reports three readings of the bank.

\begin{table}[t]
\centering
\small
\caption{The sensitivity interval under three readings of the control bank, at the registered radius. \emph{Wald} is the registered construction. \emph{Hoeffding} applies the paper's own small-sample guardrail to the control bank as well. \emph{Fixed} treats the bank as a fully measured object, so $b_C$ is a constant and the transport radius carries everything.}
\label{tab:7}
\resizebox{\ifdim\width>\textwidth \textwidth\else \width\fi}{!}{%
\begin{tabular}{llrr}
\toprule
\textbf{Reading of the control bank} & \textbf{Control half-width} & \textbf{Mean width of the $\tau$ interval} & \textbf{Pairs excluding zero} \\
\midrule
Wald (registered) & 0.057 to 0.136 & 0.335 & 0 of 10 \\
Hoeffding (own guardrail) & 0.382 to 0.468 & 0.982 & 0 of 10 \\
Fixed bank, no interval & 0 & 0.131 & 1 of 10 \\
\bottomrule
\end{tabular}}
\end{table}

The guardrail reading is nearly useless: a control half-width of 0.4 on a quantity bounded in [-1, 1] leaves an interval that could not exclude zero for any plausible audit. The fixed reading is the narrowest and it is wrong. Dropping the control interval makes one pair, SmolLM2-1.7B on ARC, exclude zero at [+0.05, +0.15], which under the claim rule would be a positive finding. That flag comes entirely from a measured control gap of -0.088 on 40 items, a number whose own interval comfortably contains zero. That leaves the registered Wald construction, which treats the bank as one realization of the item-writing procedure. The population is informal, so the interval is a heuristic and not a guarantee, and the transport radius carries the assumption across to the audited benchmark. Two of the three readings agree on all ten pairs, and the third differs only by ignoring uncertainty. What the exercise shows is that the nulls of Table 4 hold under both readings that keep the control interval.

\subsection{Baselines on the same items}
Table 8 compares CleanScore with two detectors computed on the same forms. The Clean-Eval-style statistic is the fresh-form score itself. The PaCoST-style statistic pairs, for each parent, the model's mean per-token log probability on the public form with its mean over the fresh forms (the greedy continuation for GSM8K, the chosen option for ARC) and tests the paired difference.

\begin{table}[t]
\centering
\small
\caption{Baseline detectors on the audited items. PaCoST flags a pair when the paired confidence test rejects at 0.05 with higher confidence on the public form.}
\label{tab:8}
\resizebox{\ifdim\width>\textwidth \textwidth\else \width\fi}{!}{%
\begin{tabular}{llrrrrrll}
\toprule
\textbf{Model} & \textbf{Benchmark} & \textbf{Public score} & \textbf{Fresh score} & \textbf{Confidence, public} & \textbf{Confidence, fresh} & \textbf{PaCoST p} & \textbf{PaCoST flags} & \textbf{CleanScore claim} \\
\midrule
OLMo-2-7B & GSM8K & 0.720 & 0.705 & -0.098 & -0.096 & 0.28 & no & no \\
OLMo-2-7B & ARC & 0.625 & 0.608 & -3.19 & -3.17 & 0.71 & no & no \\
SmolLM2-1.7B & GSM8K & 0.475 & 0.430 & -0.156 & -0.156 & 0.83 & no & no \\
SmolLM2-1.7B & ARC & 0.485 & 0.470 & -3.74 & -3.67 & 0.18 & no & no \\
Phi-3-mini & GSM8K & 0.820 & 0.803 & -0.119 & -0.119 & 0.94 & no & no \\
Phi-3-mini & ARC & 0.595 & 0.598 & -3.15 & -3.20 & 0.017 & \textbf{yes} & no \\
Qwen2.5-1.5B & GSM8K & 0.475 & 0.500 & -0.122 & -0.122 & 0.99 & no & no \\
Qwen2.5-1.5B & ARC & 0.495 & 0.503 & -2.82 & -2.84 & 0.68 & no & no \\
Qwen2.5-7B & GSM8K & 0.710 & 0.715 & -0.075 & -0.076 & 0.43 & no & no \\
Qwen2.5-7B & ARC & 0.590 & 0.558 & -6.09 & -6.15 & 0.49 & no & no \\
\bottomrule
\end{tabular}}
\end{table}

The detectors agree on nine of ten pairs. They disagree on Phi-3-mini and ARC. The model is measurably more confident on the public stems (paired t-test p = 0.017, Wilcoxon p = 0.010), yet its accuracy on public and fresh stems is the same: -0.003, with a Wald half-width of 0.028, or 0.192 under the registered guardrail bound. Both observations are correct. The model recognizes the surface of the public stems, and that recognition does not change which option it picks. A confidence-based detector reports the first fact and CleanScore reports the second. A verdict-style detector cannot make this distinction on its own.

No multiplicity correction was applied to the ten pairs. For the audit's own results this cannot matter, because correcting for ten tests only makes rejection harder and none of the ten was close. It matters for the detector's single flag: at p = 0.017 the Phi-3-mini result does not survive a Bonferroni threshold of 0.005 across ten pairs, and under Benjamini-Hochberg it is the smallest of ten p-values against a threshold of 0.005 and does not survive either. The flag should be read as what a detector reports before correction. Our registration did not fix a correction, and we record the omission instead of repairing it after the fact.

\subsection{What the audit could have detected}
A null is worth only as much as the audit's ability to have found something. We measure that by resampling the observed parent gaps and planting an advantage of a stated size in a random share of the parents that currently show no gap. This keeps the real variance and its heavy concentration at zero, and it moves items the way exposure to a public form moves them, from wrong or right on every form to right on the public form only. The detection rate is the share of audits that declare the advantage. We first ask this of the primary interval (Table 9) and then of the full registered claim rule (Table 10).

\begin{table}[t]
\centering
\small
\caption{Detection rate for a planted public-form advantage, 4,000 resampled audits per cell, pooling the five audited models within each benchmark.}
\label{tab:9}
\resizebox{\ifdim\width>\textwidth \textwidth\else \width\fi}{!}{%
\begin{tabular}{lrrrrr}
\toprule
\textbf{Benchmark} & \textbf{Audited items} & \textbf{delta = 0.02} & \textbf{delta = 0.05} & \textbf{delta = 0.10} & \textbf{delta = 0.20} \\
\midrule
GSM8K & 50 & 0.00 & 0.00 & 0.00 & 0.00 \\
GSM8K & 100 & 0.00 & 0.00 & 0.00 & 0.04 \\
GSM8K & 200 & 0.22 & 0.56 & 0.95 & 1.00 \\
GSM8K & 400 & 0.43 & 0.89 & 1.00 & 1.00 \\
GSM8K & 800 & 0.82 & 1.00 & 1.00 & 1.00 \\
ARC & 50 & 0.00 & 0.00 & 0.00 & 0.00 \\
ARC & 100 & 0.00 & 0.00 & 0.00 & 0.01 \\
ARC & 200 & 0.36 & 0.84 & 1.00 & 1.00 \\
ARC & 400 & 0.74 & 1.00 & 1.00 & 1.00 \\
ARC & 800 & 0.99 & 1.00 & 1.00 & 1.00 \\
\bottomrule
\end{tabular}}
\end{table}

For the primary interval, at 200 parents a ten-point advantage is detected in at least 95\% of audits and a five-point advantage about half the time on GSM8K. The minimum detectable effect is the primary half-width, from 0.035 for Phi-3-mini to 0.058 for SmolLM2-1.7B on GSM8K, and from 0.029 to 0.040 on ARC for the three pairs where the guardrail does not apply. Below 200 parents the guardrail leaves the audit with no power at all. At 100 parents it forces the distribution-free interval, whose half-width is 0.272 whatever the data show, and the detection rate is zero for every effect up to ten points and 4\% at twenty. The variance-adaptive bound of Appendix B.1 does not rescue this: at 100 parents its additive term alone is 0.207, so it is wider than the Hoeffding bound unless the sample variance is below about 0.048. An auditor limited to tens of items should not expect an informative result and should not read a wide distribution-free interval as reassurance.

The registered claim rule is much less sensitive, because it also requires the sensitivity interval to exclude zero. At the registered radius that interval has half-width 0.116 to 0.222 across the ten pairs (mean 0.167), against primary half-widths of 0.028 to 0.058 for the Wald pairs. Table 10 gives its detection rate.

\begin{table}[t]
\centering
\small
\caption{Detection rate of the full registered claim rule, which needs both the primary interval and the sensitivity interval at radius 0.02 to exclude zero, for a planted public-form advantage. 4,000 simulated audits per cell, pooling the five audited models within each benchmark, with the audited parents drawn without replacement from a frame of the benchmark's size. The first row of each block gives the primary interval alone in the same simulation, for comparison.}
\label{tab:10}
\resizebox{\ifdim\width>\textwidth \textwidth\else \width\fi}{!}{%
\begin{tabular}{lrlrrrr}
\toprule
\textbf{Benchmark} & \textbf{Audited parents} & \textbf{Control parents} & \textbf{delta = 0.05} & \textbf{delta = 0.10} & \textbf{delta = 0.15} & \textbf{delta = 0.20} \\
\midrule
GSM8K & 200 & primary interval only & 0.53 & 0.96 & 1.00 & 1.00 \\
GSM8K & 200 & 60 (registered) & 0.00 & 0.05 & 0.23 & 0.58 \\
GSM8K & 200 & 200 & 0.01 & 0.13 & 0.54 & 0.91 \\
GSM8K & 800 & 200 & 0.06 & 0.58 & 0.98 & 1.00 \\
ARC & 200 & primary interval only & 0.84 & 1.00 & 1.00 & 1.00 \\
ARC & 200 & 40 (registered) & 0.02 & 0.11 & 0.34 & 0.64 \\
ARC & 200 & 200 & 0.03 & 0.36 & 0.86 & 1.00 \\
ARC & 800 & 200 & 0.21 & 0.87 & 1.00 & 1.00 \\
\bottomrule
\end{tabular}}
\end{table}

With 200 audited parents and the registered control banks, the claim rule flags a planted ten-point advantage in 5\% of GSM8K audits and 11\% of ARC audits, against 96\% and 100\% for the primary interval in the same simulation. A 15-point advantage is flagged in 23\% and 34\%, and a 20-point advantage in 58\% and 64\%. When no advantage is planted, it flagged at most 0.4\% of audits. That holds when the benchmark's ordinary form gap equals the control's. At the edge of the transport assumption, where the benchmark's ordinary gap exceeds the control's by the radius of 0.02, the rate at the registered sizes is at most 0.1\%. The rule is therefore a strict filter against false claims and a weak detector. A control bank of 200 parents helps, but at 200 audited parents it lifts the ten-point rate only to 13\% on GSM8K and 36\% on ARC. Reaching 58\% at ten points on GSM8K takes 800 audited parents and a bank of 200.

These numbers change what the ten registered nulls mean. The primary intervals bound the observable gap to within about ten points for eight of the ten pairs, but they do not separate exposure from form mismatch. The claim rule separates them only through the control bank and the radius, and at the registered size its nulls exclude only very large advantages. Our scripts are released so an auditor can recompute these rates for another benchmark from a pilot sample. These rates are what led us to register the larger audit of Section 7.7.

\subsection{A larger audit, registered and not yet run}
Section 7.6 shows what the ten nulls above can and cannot rule out. Before running anything larger we registered the audit those numbers call for: 800 GSM8K parents against a control bank of 900 never-public problems, the same five models, a 768-token budget and a Bonferroni claim rule across the models. We have not run it. The forms are frozen and hashed. The registration makes a reading of a random sample of them by both authors a condition of the first model run, and that reading has not taken place. No result of this audit is part of the evidence of this paper. Appendix B.8 gives the design and the simulations behind it.

\section{Positive controls: how much does a paraphrase audit miss?}
Section 7.6 estimates what the audit could detect from the variance of the observed gaps. This section asks by experiment: contaminate a model with audited items, then run the same audit on it. Every study below was registered before its results were read, and before training began, except for the estimator of Study 9 (Appendix B.7). Each uses full-parameter fine-tuning on a mixture of 1,000 records (the leaked items plus background items) for three epochs and leaks the public forms of 100 randomly chosen audited parents. The other 100 parents are left unexposed and serve as a within-model control. Fresh forms never enter training, and a check confirms that no fresh-form string appears in the mixture.

Fine-tuning moves everything, and often downward, so raw before-and-after numbers mislead. Every effect below is a difference in differences: the exposed parents' change minus the unexposed parents' change on the same form type. The intervals are bootstrap intervals over parents. They capture which items were drawn and nothing else: they do not capture training variability, except where we say seeds were repeated, and they do not capture interference between trained and untrained parents. Each effect is therefore the differential change of the exposed parents under one global fine-tuning run. If training on the leaked items also moves the unexposed parents, the contrast is not the direct effect of leaking one item.

\subsection{GSM8K: four models}
Study 1 contaminated Qwen2.5-1.5B-Instruct, which is in the audited panel, so its clean audit is on record in Table 4. Leaking an item raised accuracy on the leaked wording by 24 points relative to control, a difference in differences of +0.240 with interval [+0.080, +0.400]. It raised accuracy on paraphrases the model never saw by 17.5 points, +0.175 with interval [+0.035, +0.315]. Only the difference, 6.5 points, appears as a public-form advantage. The transferred share is 73\%, with a wide interval of [19\%, 168\%] because it divides one noisy quantity by another. The unexposed hundred behave as a control should: fine-tuning damaged general arithmetic, lowering accuracy by nine points on both the public and the fresh forms of untouched parents, and the measurable gap moved by 0.000 (Appendix B.6).

To see whether this is one model's habit, we repeated it on SmolLM2-1.7B-Instruct (Study 2) and, in a separately registered Study 3, on OLMo-2-1B-Instruct and StableLM-2-1.6B-Chat. Each of those registrations allowed one training run, so the first four-model result rested on a single run per model. Study 9 reran all four models under three further training seeds, with the seeds fixed before the first run and the pooling and the reporting rules fixed before any result was read (Appendix B.7). The falsifier registered for Study 3 still applies: two or more models with a fresh-form effect at or below zero alongside a clearly positive public-form effect would mean the transfer is not general. Table 11 gives each model's mean over its four runs.

\begin{table}[t]
\centering
\small
\caption{Four contaminated models from four families, each trained four times. Difference-in-differences against the unexposed control parents of the same model, as the mean over the model's four runs, with the registered two-level 95\% intervals. The visible gap is what a paraphrase audit observes. The last column is the range of a model's four runs, on the leaked wording and on unseen paraphrases.}
\label{tab:11}
\resizebox{\ifdim\width>\textwidth \textwidth\else \width\fi}{!}{%
\begin{tabular}{llllrl}
\toprule
\textbf{Model} & \textbf{Family} & \textbf{DiD public} & \textbf{DiD fresh} & \textbf{Visible gap} & \textbf{Range over seeds} \\
\midrule
Qwen2.5-1.5B & Qwen & +0.243 [+0.099, +0.387] & +0.154 [-0.0003, +0.310] & +0.089 & 0.08; 0.16 \\
SmolLM2-1.7B & SmolLM2 & +0.125 [-0.016, +0.267] & +0.040 [-0.043, +0.123] & +0.085 & 0.08; 0.01 \\
OLMo-2-1B & OLMo & +0.460 [+0.316, +0.607] & +0.361 [+0.230, +0.494] & +0.099 & 0.08; 0.12 \\
StableLM-2-1.6B & StableLM & +0.180 [+0.038, +0.324] & +0.213 [+0.074, +0.350] & -0.033 & 0.09; 0.15 \\
\textbf{Pooled} &  & \textbf{+0.252 [+0.180, +0.325]} & \textbf{+0.192 [+0.125, +0.260]} & +0.060 &  \\
\bottomrule
\end{tabular}}
\end{table}

All four models show a positive mean effect on paraphrases they never saw, and no single run of any model has one at or below zero. On its own, two of the four intervals exclude zero. The interval for Qwen2.5-1.5B just touches it (its lower end is -0.0003), and SmolLM2, which absorbed the leak least, includes zero comfortably; the fresh-form effect is positive in 97\% and 82\% of their draws. The falsifier did not fire.

Pooled across four models and four seeds (a descriptive average, since the models differ), the effect is +0.252 on the memorized wording and +0.192 on unseen paraphrases, and the pooled transferred share is 76\% with interval [50\%, 113\%]. The interval comes from the registered two-level resampling, which draws the parents once for every model and seed and adds a seed component estimated from each model's four runs. The share exceeds one half in 97\% of draws, but its lower limit is just under one half, so we say that a substantial share of what a model gains from seeing an item does not stay attached to the wording it saw, and not that most of it does. The share is not bounded away from unity: the interval runs to 113\%, so the data are consistent with all of the effect transferring and none of it being visible to a paraphrase audit. The registered interval draws the seed component of the public and fresh effects independently. Drawing it as one shared variable, the covariance-aware version, gives a narrower interval, [54\%, 105\%], around the same 76\%. One of 20,000 draws had a pooled public effect at or below 0.05 and was left out of the share. The three extra seeds moved the point estimate by one point, from 77\% to 76\%, and did not narrow the interval, because the same parents enter every run and no number of seeds averages that sampling error away.

These results come from four models of 1B to 1.7B parameters, each fine-tuned four times with 100 leaked items, and the leak was deliberate and concentrated. They show that transfer occurs and is substantial in this setting. They do not show its size in the wild.

\begin{figure}[t]
\centering
\includegraphics[width=\textwidth]{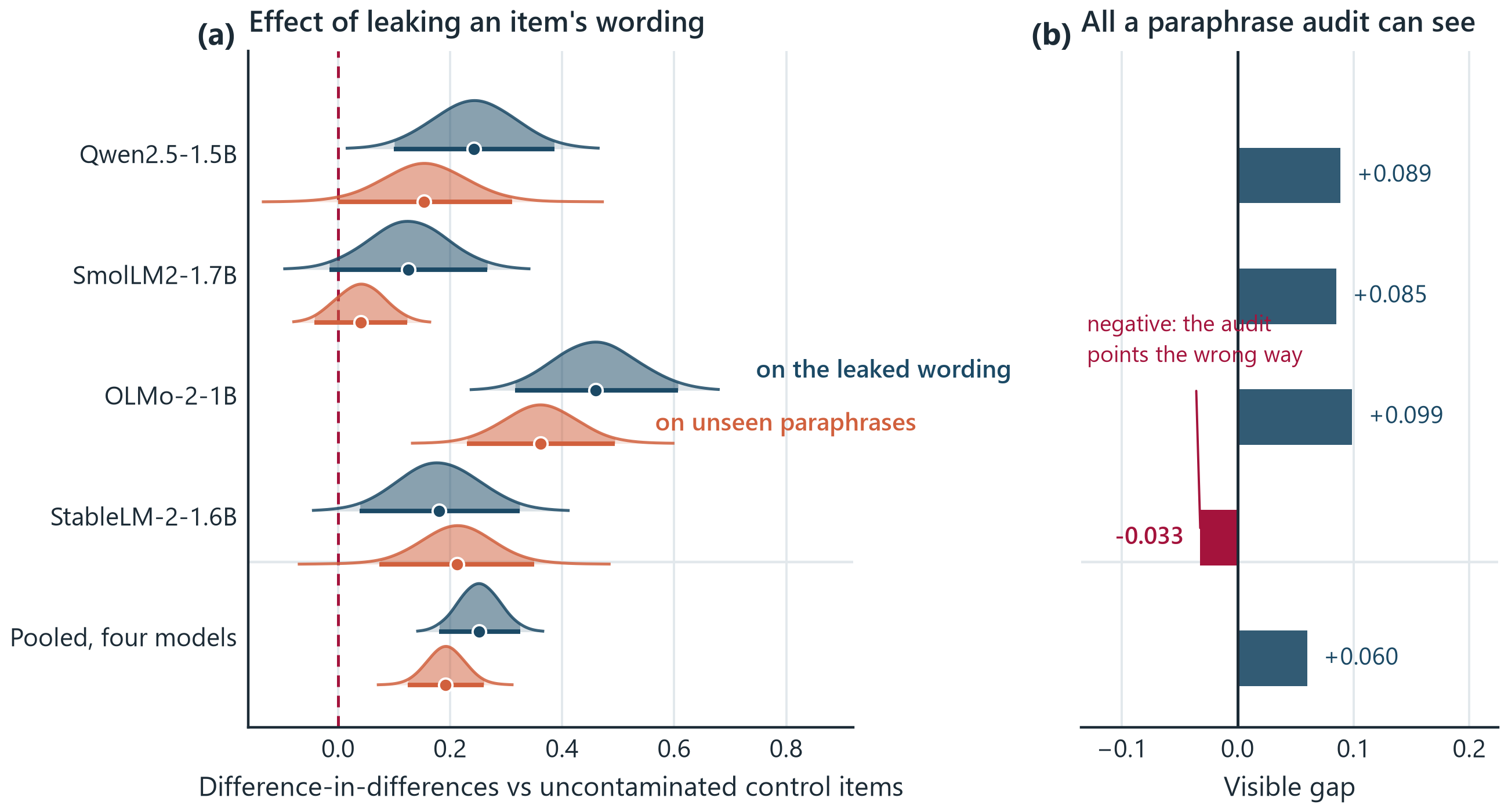}
\caption{Left: for each model, the effect of leaking an item on the wording trained on and on paraphrases never seen, as differences in differences against uncontaminated control items, averaged over four training seeds, with the registered two-level intervals. Right: the visible gap, the difference between the two, which is all a paraphrase audit observes. For StableLM-2-1.6B it is negative.}
\label{fig:4}
\end{figure}

Two features of Figure 4 need stating. The effect varies across models, from 12.5 points to 46 on the memorized wording. OLMo-2-1B lost 34 points on its untouched control parents in its registered run, so much of its +0.460 is protection of the items it saw from a degradation that hit everything else. That is still item-specific knowledge, but it is a different regime from Qwen's nine-point drift, and the two are not measurements of the same quantity.

The visible gap is also not stable: +0.089, +0.085 and +0.099 for the first three models, then -0.033. StableLM is the sharpest case. Leaking a hundred items raised its accuracy on them by eighteen points relative to control, yet a paraphrase audit of it would observe a gap of -0.033, negative in three of its four runs and pointing the wrong way, and would report no evidence of contamination while nearly a fifth of the leaked items had been learned.

Training variability is large next to these gaps. The range across a model's four runs is 8 to 9 points on the leaked wording and 1 to 16 points on unseen paraphrases, which is why the pooled interval carries a seed component.

The registered claim rule did not fire on the Qwen model, and the instrument behaved as Section 7.6 predicts. Averaged over the whole bank, where only half the parents were contaminated, the audit's estimate moved from -0.025 to +0.008, a shift of +0.033 against a minimum detectable effect of 0.057 for this model at 200 parents. The exact test gives p = 0.45. Restricted to the exposed hundred, the gap of +0.040 is not significant either (p = 0.23).

\subsection{ARC: a leak that failed and one that worked}
GSM8K is scored by letting the model write an answer. ARC is scored differently: nothing is generated, and each option is ranked by how likely its text is as a continuation of the question. We registered a second positive control to see whether the finding survives that change. It needed an extra form. The released ARC fresh forms rewrite the stem and keep the options word for word, so a model that memorized only an option list would score the same on the public and fresh forms. We therefore wrote a third form for each exposed item, a deep paraphrase that rewords the stem and the options together while keeping what each option refers to and where the correct one sits. The model was StableLM-2-1.6B.

Study 4 failed. Fine-tuning on the leaked items lowered accuracy on the leaked wording by 18 points, while untouched items in the same model rose by four. The registered manipulation check, the difference in differences on the trained wording, was -0.220 against a required floor of +0.05 (Appendix B.5). By the rule fixed in advance this ends the study. No contamination was planted, so the study says nothing about whether contamination transfers on ARC, in either direction, and the transfer shares are withheld.

The cause was a mismatch between what the leak planted and what the scorer reads. The exposure record ended with the line ``Answer: A.'', the form a leaked multiple-choice item takes when written down. The scorer never shows the model an option list. It asks which option's text is the likelier continuation. Training installed a pull toward the first-listed option, right a quarter of the time. On trained items the model picked the first-listed option 68\% of the time. On the released fresh forms, which keep the option strings, it was 62\% and 60\%. On the deep paraphrases, where those strings are gone, it fell to 33\%, and on never-seen items it stayed at 19\%.

Training loss fell from 1.01 to 0.16 as if all were well. Only the numerical criterion fixed beforehand caught the failure. A positive control must leak the item in a form the scoring path can reward. One more deviation: the registration made a human reading of every deep-paraphrase item a precondition, and the run went ahead before that reading was finished. This does not touch the outcome, because the manipulation check uses the original ARC items, but we record it.

Study 5 changed one thing and was run once. The exposure record became the question followed by the text of the correct option, with no option list and no letter, which is the form a scraped question-and-answer pair takes and the form the scorer reads. The model, the exposed parents, the mixture, the schedule and the seed were held fixed. The manipulation check moved from -0.220 to +0.490, which confirms the diagnosis. Leaking an item raised accuracy on the leaked wording by 49 points and on the rewritten stems by 51 points, so the gap an auditor would compute is -0.020, with an interval from -0.110 to +0.070. The model carries 49 points of deliberate contamination and the ARC audit of Section 7 sees none of it. The accuracies are near perfect: the exposed model answered all 100 trained items correctly and 98 of 100 rewritten ones. The blindness is therefore mechanical, since a memorized answer string sits in both the original and the rewrite. Any ARC-style audit whose rewrites hold the options fixed is blind to a leak of this form.

Study 5 is a further ARC study, and we owe the reader the reasoning. Study 4's stopping rule said no further ARC studies whatever the outcome. We ran Study 5 because the fault lay in the instrument and not in the result, which is provable from the two scripts without running either. Study 4 stays in the paper at full length, and every interpretation of Study 5 was fixed before any number existed. A reader who does not accept this should treat Study 4 as the result and Study 5 as an addendum to discount.

\subsection{ARC: the complete deep-paraphrase bank, a null run, and four seeds}
In Study 5 the deep-paraphrase arm was the one built to see past the memorized string, and it was demoted by its own registered check. Its counterfactual was borrowed from the control group's change on fresh forms, because only 25 of the 100 control parents had a deep paraphrase, and the assumption behind the borrowing was checked on 20 parents against a tolerance of 0.05. The check returned -0.100, with a bootstrap interval from -0.350 to +0.150. Twenty parents cannot resolve that tolerance, so the arm was reported as descriptive only.

The remedy is to remove the need for the assumption. We wrote deep paraphrases for the 75 control parents that lacked one, under the six drafting rules now in force, and reran the experiment (Study 6), so that the deep-paraphrase effect is the same quantity as the other two: the exposed group's change minus the control group's change on the same form type. The public and fresh arms reproduced to four decimal places, which confirms the pipeline is deterministic and is not evidence about sampling variability.

The bank holds 200 items, of which 24 have options that cannot be reworded (balanced equations, bare quantities, proper nouns). They are excluded from the deep-paraphrase arm, which is why it has 93 exposed and 83 control parents. Table 12 gives all three arms.

\begin{table}[t]
\centering
\small
\caption{The complete deep-paraphrase bank. Accuracy on the leaked parents beside the untouched ones, for all three form types, before and after the same fine-tuning as Study 5. The deep-paraphrase row is now a direct comparison rather than a substituted one.}
\label{tab:12}
\resizebox{\ifdim\width>\textwidth \textwidth\else \width\fi}{!}{%
\begin{tabular}{llrrrl}
\toprule
\textbf{Parents} & \textbf{Form} & \textbf{Before} & \textbf{After} & \textbf{Difference in differences} & \textbf{95\% CI} \\
\midrule
Exposed (n = 100) & public, trained on & 0.460 & 1.000 & \textbf{+0.490} & [+0.360, +0.620] \\
Exposed (n = 100) & fresh, options kept & 0.420 & 0.980 & \textbf{+0.510} & [+0.395, +0.625] \\
Exposed (n = 93) & deep paraphrase & 0.376 & 0.602 & \textbf{+0.238} & [+0.089, +0.387] \\
Unexposed (n = 100) & public & 0.400 & 0.450 & — & — \\
Unexposed (n = 100) & fresh & 0.400 & 0.450 & — & — \\
Unexposed (n = 83) & deep paraphrase & 0.410 & 0.398 & — & — \\
\bottomrule
\end{tabular}}
\end{table}

Leaking an item raised accuracy by 24 points on a rewrite that changes the stem and all four options, against control parents that moved by -1 point on the same kind of rewrite, a difference in differences of +0.238 with interval [+0.089, +0.387]. No option string from a deep paraphrase appears in training. So contamination on ARC reaches wording the model never saw. The transfer found on GSM8K in Section 8.1 also occurs on a second benchmark, under a different scoring path, in the one model tested.

Three qualifications go with it. The earlier demotion was justified: on 83 control parents, fine-tuning moved fresh forms up by 5 points and deep paraphrases down by 1, an asymmetry of about -0.062 in the direction the small check reported, and the borrowed counterfactual had biased the estimate downward, +0.176 against the direct +0.238. The ceiling matters: accuracy on trained items is exactly 1.000, so the public arm is censored at 49 points or more, and we report differences in differences and no share. And the audit still sees none of it: the gap between public and released fresh forms is -0.020, with 49 points of contamination present and about half of the effect surviving a total rewrite. Figure 5 shows the three arms.

\begin{figure}[t]
\centering
\includegraphics[width=\textwidth]{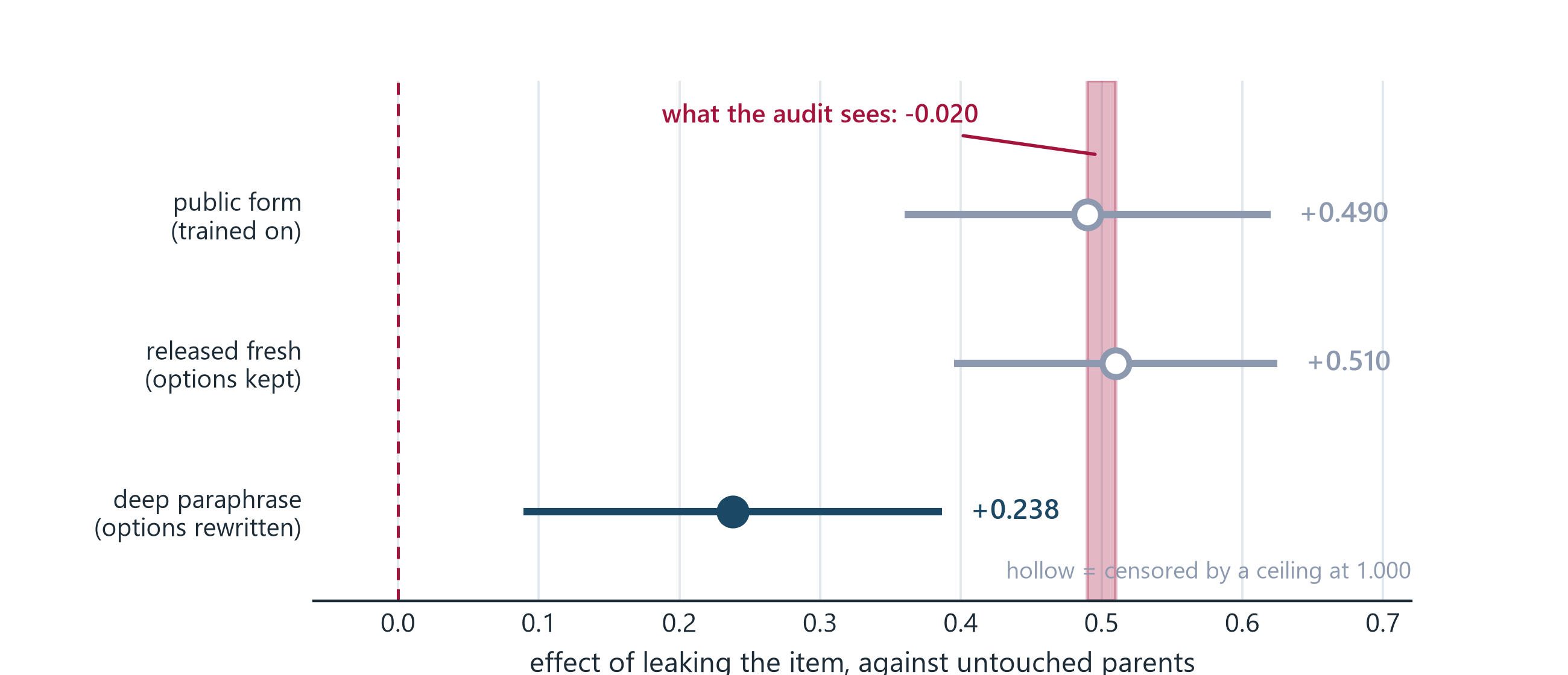}
\caption{The three arms of the complete-bank study with bootstrap intervals over parents. The public and fresh-form effects are censored by a ceiling at 1.000 and are shown hollow. The deep-paraphrase effect is not censored, rests on a direct comparison against control parents that carry the same form type, and excludes zero. The bracket marks the quantity a paraphrase audit of this benchmark reports.}
\label{fig:5}
\end{figure}

Two further experiments address training variability. Study 7 is a negative control on the procedure. We ran the same fine-tuning with no leak: all 1,000 training records are background items, and the script asserts that no audited wording appears. Table 13 gives the result. The largest difference in differences is 0.020, inside the 0.05 the registration set for calling the procedure clean, and all three intervals contain zero. The planted effects are twelve to twenty-five times the empirical null, so fine-tuning alone does not manufacture them. This is a single run, so the null carries its own unmeasured seed variability, but a single draw within two points of zero is weak evidence that the null is near zero, and a draw near fifty would have meant the design was broken.

\begin{table}[t]
\centering
\small
\caption{The same statistic with the leak removed. No audited item's wording is in the training mixture, so every entry in the last column is an empirical null for the whole procedure, and the planted effects from Study 6 are shown beside them for scale.}
\label{tab:13}
\resizebox{\ifdim\width>\textwidth \textwidth\else \width\fi}{!}{%
\begin{tabular}{llrlr}
\toprule
\textbf{Form family} & \textbf{Accuracy on leaked-half parents} & \textbf{Difference in differences, no leak} & \textbf{95\% CI} & \textbf{Study 6, with the leak} \\
\midrule
Public form & 0.460 to 0.460 & -0.010 & [-0.120, +0.100] & \textbf{+0.490} \\
Released fresh & 0.420 to 0.435 & +0.020 & [-0.060, +0.100] & \textbf{+0.510} \\
Deep paraphrase & 0.376 to 0.344 & -0.008 & [-0.147, +0.128] & \textbf{+0.238} \\
\bottomrule
\end{tabular}}
\end{table}

Study 8 repeats the leaked experiment of Study 6 under three further training seeds. Every earlier registration forbade extra seeds, a rule written to stop an experiment being repeated until it gave a convenient answer. We ran them anyway as an authors' override recorded in the registration and the deviation log, because a replication that can only weaken a published estimate does not do what the rule is for. The binding commitment, fixed before the runs, is that every seed is reported and none dropped. The same hundred parents are leaked and the same records trained on in each run; only the trainer seed and the record order change. Table 14 gives every run.

\begin{table}[t]
\centering
\small
\caption{All four training seeds, including Study 6's. Each row is a complete rerun of Section 8.3 differing only in the trainer seed and the order of the training records.}
\label{tab:14}
\resizebox{\ifdim\width>\textwidth \textwidth\else \width\fi}{!}{%
\begin{tabular}{lrrr}
\toprule
\textbf{Training seed} & \textbf{Public form} & \textbf{Released fresh} & \textbf{Deep paraphrase} \\
\midrule
20260824 (the registered run, Section 8.3) & +0.490 & +0.510 & \textbf{+0.238} \\
20260828 & +0.460 & +0.510 & +0.179 \\
20260829 & +0.500 & +0.495 & +0.204 \\
20260830 & +0.480 & +0.505 & +0.195 \\
\textbf{Mean} & \textbf{+0.483} & \textbf{+0.505} & \textbf{+0.204} \\
Spread & 0.040 & 0.015 & 0.059 \\
\bottomrule
\end{tabular}}
\end{table}

All four manipulations pass, and the deep-paraphrase effect is positive on all four. The public-form spread of 0.040 is inside the 0.10 tolerance fixed in advance, so optimization variability is small relative to the planted effect. The registered run landed at the favorable end: its +0.238 is the largest deep-paraphrase estimate of the four, the range runs down to +0.179, and the four-seed mean is +0.204. The seed was fixed before any of this and the other three were added afterward, but a reader comparing the single-seed figure with the mean should know where the registered run fell. We therefore quote the deep-paraphrase effect as about 20 points, with a four-seed range of 18 to 24, and keep +0.238 as the registered single-seed result. The ordering of the three arms holds on every seed, and the paper's argument rests on it: public and fresh effects are large and nearly equal, so the audit's gap stays near zero, and the deep-paraphrase effect is smaller and positive on every seed. Four seeds of one model at one dose on one benchmark bound optimization variability in that setting only.

\section{Limitations}
\textbf{What the estimand is.} CleanScore cannot show that a named model trained on a named item, and a null does not show that an audited model did not. The estimand is the surface-form advantage: fresh forms keep every number, fact and answer and change only the wording. A model can depend on the training distribution of a benchmark, the pattern of problem types and answer magnitudes, without depending on the wording of any item. Studies that write new problems in the style of GSM8K measure that broader dependence and find it for some model families (Zhang, Da, et al., 2024). Our nulls are consistent with such effects being present in the models we audited.

\textbf{The forms.} A language model drafted the fresh forms and the private control banks under written rules, and every form passed a deterministic check. The authors' reading, the second check, came after the results and was prioritized: it covered a released packet and not all 1,500 forms. A reading that follows the results cannot have shaped them, but it cannot rule out a defect among the forms it did not reach. The drafting model is from a different provider and family than every audited model, but a style shared across the forms would pass the mechanical check. 

\textbf{The deep-paraphrase bank.} The bank behind Sections 8.2 and 8.3 is in a weaker state. All 200 items were screened against their public originals by a language model in two passes, with findings released with the paper. No person has read any of them, and the reading that Study 4's registration made a precondition has not been done. Machine screening is not a substitute: the same model family drafted the items and shares whatever blind spot put a fault in them. The mechanical checks cannot cover the semantic rule that rewording an option must not make a distractor correct, and the deep-paraphrase accuracies depend on it directly. A defect there would move only the deep-paraphrase arm, and because the estimate is a difference in differences it would have to fall unevenly between exposed and control halves to bias it. 

Three items flagged by the screening and the 21 near-copy GSM8K forms (Appendix B.3) were left in the banks, because removing items after seeing results is what makes a bank untrustworthy. Dropping the three flagged items moves the deep-paraphrase estimate by between +0.003 and +0.005 across the four seeds, toward a larger effect.

\textbf{The audit.} The registered 320-token budget truncated up to 40.7\% of GSM8K forms. Amendment 1 (two Qwen models) and Amendment 2 (the other three) rerun GSM8K at 768 tokens, and the nulls survive. SmolLM2-1.7B still has 2.0\% of its audit forms truncated, above the 1\% that the amendment allows, so its rerun is partial (Section 7.3). The control banks are small and composed without a frame, so the interval on them is a heuristic (Section 7.4), and the transport radius $\rho^*$ = 0.02 is a judgment reported beside the full grid. The audit is powered for large effects: the primary interval finds a ten-point advantage in at least 95\% of audits, but the registered claim rule finds one in only 5\% to 11\% of audits (Section 7.6). The ten nulls bound large surface-form inflation and say little about small inflation. The ARC arm is blind by construction to memorization attached to the options, and it is scored from option log probabilities, so it is gray-box in practice. The generated-letter readout of Amendment 2 needs generated text only and gives estimates within 0.010 of zero, but four of its five primary intervals are the distribution-free bound at half-width 0.192, and two of its control banks show a gap of +0.088. No sampling frame defines the composed control banks, so the claim rule has no demonstrated false-claim rate for benign rewrites in general. The panel is five open models between 1.5B and 7B parameters on two benchmarks, and it says nothing about frontier or closed models. The controlled-exposure validation used 0.5B and 360M models, and its registered model is direction-consistent but underpowered.

\textbf{The registered larger audit.} It has not been run, so none of its results is evidence here (Section 7.7 and Appendix B.8). Its control bank is written by the same language model that wrote the fresh forms, so its gap would compare model-written with model-rewritten text, while the benchmark's gap compares human-written originals with model rewrites, and the transport radius carries the whole difference. A human-written control stratum was registered as optional and has not been written. Only the two authors would read its forms, and an outside reader on a sample would strengthen that.

\textbf{The positive controls.} The GSM8K estimate rests on four models of 1B to 1.7B parameters, one fine-tuning regime, 100 leaked items each and four training runs per model. Two of the four are individually significant on unseen paraphrases and the pooled estimate is +0.192, with a 95\% interval from +0.125 to +0.260, but the models are small, the leak was deliberate, and the effect varies fourfold across models. OLMo-2-1B's estimate reflects protection from forgetting as much as gain. Pooling puts the transferred share at 76\%, with an interval of [50\%, 113\%]. Narrowing it would take more parents and more models, not more seeds, because every run shares the same 200 parents. Three earlier attempts at this experiment failed to implant any contamination; they are in the deviation log and contribute nothing to the result. 

Every interval in Section 8 is a bootstrap over parents. It does not capture other effects of a fine-tuning run or interference between trained and untrained parents, and it captures training variability only where seeds were repeated: through the seed component of the GSM8K pooled interval, and in the four-seed ARC replication, which covers one model at one dose. A registration that forbids extra seeds also prevents replication, and a registration that wants both should say in advance how many seeds it will run and commit to reporting all of them. ARC rests on one model of 1.6B parameters under one procedure, so it shows that transfer exists on this scoring path and not how large it is in general.

\section{Discussion and conclusion}
Benchmark contamination is partly a data-access problem and partly an identification problem. Black-box output differences can show that a public score is fragile, but they cannot by themselves say why. CleanScore is built around that boundary: it reports an observable gap, a private-control estimate, and the range of conclusions a stated transport radius allows.

The experiments support three conclusions. First, the controlled-exposure validation found what it was built to find in one of two families and nothing it was built to ignore in either: planted public forms raised the public-form advantage, planted fresh forms did not, and the effect spread across the bank as the lock predicted.

Second, the real audit found no exposure-consistent advantage under a rule fixed in advance, and the result survives two pre-declared reruns, one that removes an answer-length artifact and one that scores ARC from generated letters. The upper end of the interval is about five points or less for four of ten pairs and at most 21 points for all ten. A confidence-based detector's single flag, on a pair with a zero accuracy gap, shows why confidence and score must be reported separately.

Third, the audit's precision is limited by the control bank, and the registered claim rule is far less sensitive than the interval it contains. We recommend that an audit state the detection rate of its own claim rule, build a control bank at least as large as the audit sample, register a multiplicity correction, and use the guardrail bound in both components of the sensitivity interval. A larger audit designed to follow these recommendations is registered (Section 7.7 and Appendix B.8) and has not been run.

The positive controls say what the nulls leave out. When we contaminated four models ourselves and could see the ground truth, a substantial share of what a model gained from an item survived rewording: pooled over four models and four seeds, +0.192 on unseen paraphrases against +0.252 on the wording trained on. Between about half and all of the effect never reaches a paraphrase audit. On one model the visible gap was negative while nearly a fifth of the leaked items had been learned. On ARC a 49-point leak left the audit gap at -0.020, and about 20 points survived a rewrite of the stem and all four options, averaged over four seeds. A paraphrase-based detector, ours included, measures only the part of contamination that stays attached to the surface. The null results are therefore not a claim that the audited models are clean. They state that, for these benchmarks and models, rewriting the surface of a question does not change the answer within stated bounds.

CleanScore's purpose is not to turn a gap into an accusation. It says what the data show, what depends on an assumption, and how much that assumption matters.

\textbf{Data and code.} The registrations (with the name of the form-drafting model redacted), sampling scripts and seeds, frozen parent lists, item banks, item-verification script, a log of every deviation from a registration, model revisions, per-form outputs with log probabilities, and the simulation, analysis and plotting code are in a public repository, https://github.com/jopoku16/cleanscore-audit, released as version v1.0.0 with a SHA-256 digest for every file. Driver logs of the model runs and the notebooks that launched them are not included; the numeric outputs of the runs are. The fresh forms and control banks were kept private until scoring was complete. The deviation log records when the authors' reading of the forms took place relative to the results. The code is under the MIT license and the data under CC BY-SA 4.0. The authors maintain the repository and answer questions through its issue tracker, and a DOI will be created from the release at acceptance.

\section{Declarations}
\textbf{Author contributions.} Jeffery Opoku wrote the manuscript. David Banahene helped run and analyze some of the data.

\textbf{Use of AI tools.} Claude,an AI assistant from Anthropic, was used to help debug a few errors in the code for this work. The
authors take full responsibility for the content of this paper.

\textbf{Ethics.} The study involves no human participants and no personal data. The only human judgments are the two authors' own readings of benchmark text.

\impact{Benchmark scores steer model choice, funding and public claims, so a way to show how fragile a score is can improve those decisions. CleanScore needs only a correct-or-incorrect score for each question form and no training data, which lets outside parties run it on models they do not control. Reading its output carries three risks. A null result bounds surface-form inflation only: Section 8 shows that a substantial share of a planted contamination effect is invisible to a paraphrase audit, so a clean result must not be quoted as proof that a model is uncontaminated. A positive result does not show that a named developer trained on a benchmark. A developer could also train a model on paraphrases and pass the audit. Releasing the forms lets others repeat the audit, and it lets later models train on them, so the released forms should be treated as used. We recommend that any report of an audit give the interval and the transport radius.}

\acks{This work received no external funding. The authors declare no competing interests. Model runs used free GPU sessions provided by Kaggle.}

\appendix
\section{Mathematical core}

\subsection{Observable target}
Fix a model, prompt, decoding rule, and scoring rule. Benchmark parent item $i$ has one public original form and $K$ independently written fresh forms. Let $Y_{iO}$ be correctness on the original form and let $Y_{iF1},\ldots,Y_{iFK}$ be correctness on the fresh forms. Each outcome lies in ${0,1}$.
Define the parent-level public-form advantage
\begin{equation*}
D_i=Y_{iO}-\frac{1}{K}\sum_{k=1}^K Y_{iFk}.
\end{equation*}

Thus $D_i\in[-1,1]$. For a finite benchmark containing $N$ parent items, the audit target is
\begin{equation*}
\Delta_N=\frac{1}{N}\sum_{i=1}^N D_i.
\end{equation*}

This is an observable score gap. It is not, by itself, a contamination probability or proof that the model saw the benchmark during training.
If decoding is stochastic, all repetitions for one parent item remain inside the same parent cluster. They must not be treated as independent benchmark items.

\subsection{Conservative interval for a sampled audit}
Suppose $n$ parent items are selected uniformly without replacement from the $N$-item benchmark. Let
\begin{equation*}
\widehat\Delta=\frac{1}{n}\sum_{i\in S}D_i.
\end{equation*}

\medskip\noindent\textbf{Proposition 1.}\ \itshape
For any $\alpha\in(0,1)$,
\begin{equation*}
\Pr\left(
|\widehat\Delta-\Delta_N|
\leq
\sqrt{\frac{2\log(2/\alpha)}{n}}
\right)\geq 1-\alpha.
\end{equation*}

\upshape\medskip\noindent\textit{Proof.}
The sampled values lie in an interval of width two. Sampling without replacement is no less concentrated than independent sampling with replacement from the same finite population. Applying Hoeffding's inequality to the with-replacement comparison gives
\begin{equation*}
\Pr(|\widehat\Delta-\Delta_N|\geq\varepsilon)
\leq 2\exp(-n\varepsilon^2/2).
\end{equation*}

Solving for $\varepsilon$ proves the claim. The interval is conservative. If every benchmark item is audited, there is no item-sampling uncertainty for the finite-benchmark mean, although model-output randomness may still remain.

\subsection{Stratified audit}
Real audits should sample separately by domain and difficulty. Let stratum $h$ contain $N_h$ items, let $W_h=N_h/N$, and sample $n_h$ items without replacement. Define
\begin{equation*}
\widehat\Delta_{\mathrm{str}}=
\sum_{h=1}^H W_h\widehat\Delta_h.
\end{equation*}

\medskip\noindent\textbf{Proposition 2.}\ \itshape
If sampling is independent across strata, then
\begin{equation*}
\Pr\left(
|\widehat\Delta_{\mathrm{str}}-\Delta_N|
\leq
\sqrt{
2\log(2/\alpha)
\sum_{h=1}^H\frac{W_h^2}{n_h}
}
\right)
\geq 1-\alpha.
\end{equation*}

\upshape\medskip\noindent\textit{Proof.}
Each sampled value in stratum $h$ enters the weighted estimator with coefficient $W_h/n_h$. Its weighted range has width $2W_h/n_h$. Hoeffding's inequality for the weighted sum therefore has total squared range
\begin{equation*}
\sum_{h=1}^H n_h\left(\frac{2W_h}{n_h}\right)^2
=4\sum_{h=1}^H\frac{W_h^2}{n_h}.
\end{equation*}

The same comparison applies to sampling without replacement inside each stratum. Substitution into the two-sided Hoeffding bound proves the result.
This proposition gives a transparent budget rule: more items should be assigned to strata with large benchmark weight and large observed variation. A variance-adaptive allocation may be used as a secondary design, but it must be fixed using pilot data rather than the final audit outcomes.

\subsection{Practical design-based precision interval}
The distribution-free interval above can be very wide. The main paper therefore reports a second, standard survey-sampling interval. Let $s_h^2$ be the sample variance of the $D_i$ values in stratum $h$, and let $f_h=n_h/N_h$. Then
\begin{equation*}
\widehat V=
\sum_{h=1}^H W_h^2(1-f_h)\frac{s_h^2}{n_h}
\end{equation*}

is unbiased for the design variance of the stratified sample mean under independent simple random sampling within strata. Under the usual finite-population central limit conditions,
\begin{equation*}
\widehat\Delta_{\mathrm{str}}
\pm z_{1-\alpha/2}\sqrt{\widehat V}
\end{equation*}

has asymptotic $1-\alpha$ coverage. This interval is a standard precision tool, not a new theorem. CleanScore reports it beside the conservative bound and checks its coverage in simulation. The two intervals must not be described as having the same guarantee.
The extended simulation shows that the approximation can under-cover when the audit is small and most parent-level gaps are zero. An independent exact calculation enumerated all possible sample compositions under four discrete finite populations. In the sparse case with population size 1,000 and sample size 50, exact Wald coverage was 0.9127; it rose to 0.9467 at sample size 200 and 0.9499 at sample size 500. This confirms that the small-sample failure is not Monte Carlo noise. CleanScore therefore uses the following registered guardrail: if the audit contains fewer than 200 parent items, or fewer than 30 sampled parent items have a nonzero gap, the distribution-free interval is the primary inferential result and the Wald interval is descriptive. These thresholds are operating rules supported by stress tests, not universal mathematical cutoffs.

\subsection{Why exposure is not identified from a score gap}
Let $b_B$ be the original-versus-fresh gap that the same model would have shown on benchmark $B$ without prior exposure. Let $\tau_B$ be the exposure-related part of the gap. We define
\begin{equation*}
\Delta_B=\tau_B+b_B.
\end{equation*}

The decomposition is a definition, not an identification result. The observed audit identifies $\Delta_B$, but it does not identify $\tau_B$ unless something is learned or assumed about the model-specific counterfactual gap $b_B$. Human judgments of question difficulty alone do not guarantee a bound on a model's accuracy gap.

\subsection{Negative-control calibration with a transport radius}
Build a private control bank $C$ that was never public. For every control parent item, create one public-like first form and $K$ later fresh forms using the same writing and review protocol as the audited benchmark. Because the whole bank remains private until querying is complete, its average form gap $b_C$ can be estimated without known item exposure.
To use this control for benchmark $B$, state a transport assumption
\begin{equation*}
|b_B-b_C|\leq\rho,
\end{equation*}

where $\rho\geq0$ is reported across a sensitivity grid.

\medskip\noindent\textbf{Proposition 3.}\ \itshape
Let $I_\Delta=[L_\Delta,U_\Delta]$ cover $\Delta_B$ with probability at least $1-\alpha_\Delta$, and let $I_C=[L_C,U_C]$ cover $b_C$ with probability at least $1-\alpha_C$. Under the stated transport assumption,
\begin{equation*}
I_\tau(\rho)=
[L_\Delta-U_C-\rho,\;U_\Delta-L_C+\rho]
\end{equation*}

covers $\tau_B$ with probability at least $1-\alpha_\Delta-\alpha_C$.
The statement is exact only when both component intervals have their stated coverage. In the audit the $\Delta$ component is a Wald interval and the control component is an interval on a composed bank with no sampling frame (Section 7.4), so for the audit the guarantee is approximate.

\upshape\medskip\noindent\textit{Proof.}
On the event that both confidence intervals cover, $b_B\in[L_C-\rho,U_C+\rho]$. Since $\tau_B=\Delta_B-b_B$, subtracting the widest allowed interval for $b_B$ from the interval for $\Delta_B$ gives the stated endpoints. The union bound shows that the joint coverage event has probability at least $1-\alpha_\Delta-\alpha_C$.
The value of $\rho$ cannot be hidden or selected because it produces a favorable conclusion. The paper must show the full sensitivity curve and identify the smallest $\rho$ at which the practical conclusion changes.

\subsection{Proposition 3b}
Let the audit parents and the control parents be independent samples. Let $\widehat V_\Delta$ be the finite-population variance estimate of $\widehat\Delta$ and $\widehat V_C$ the variance estimate of $\widehat b_C$, and let $z=z_{1-\alpha/2}$. Define
\begin{equation*}
I^{\mathrm{ind}}_\tau(\rho)=
\Bigl[\widehat\Delta-\widehat b_C-z\sqrt{\widehat V_\Delta+\widehat V_C}-\rho,\;
\widehat\Delta-\widehat b_C+z\sqrt{\widehat V_\Delta+\widehat V_C}+\rho\Bigr].
\end{equation*}

If $|b_B-b_C|\le\rho$ and $\widehat\Delta-\widehat b_C$ is approximately normal around $\Delta_B-b_C$ with variance $V_\Delta+V_C$, then $I^{\mathrm{ind}}_\tau(\rho)$ covers $\tau_B$ with probability at least $1-\alpha$, approximately.

\upshape\medskip\noindent\textit{Proof.}
Because the two samples are independent, $\operatorname{Var}(\widehat\Delta-\widehat b_C)=V_\Delta+V_C$. Under the approximation, the event $|(\widehat\Delta-\widehat b_C)-(\Delta_B-b_C)|\le z\sqrt{\widehat V_\Delta+\widehat V_C}$ has probability about $1-\alpha$. Since $\tau_B=\Delta_B-b_B=(\Delta_B-b_C)-(b_B-b_C)$ and $|b_B-b_C|\le\rho$, we have $\tau_B\in[\Delta_B-b_C-\rho,\;\Delta_B-b_C+\rho]$. Write $c=\Delta_B-b_C$, $\widehat c=\widehat\Delta-\widehat b_C$ and $h=z\sqrt{\widehat V_\Delta+\widehat V_C}$. On the event $|\widehat c-c|\le h$ we have $\widehat c-h\le c\le\widehat c+h$, so $[c-\rho,c+\rho]\subseteq[\widehat c-h-\rho,\widehat c+h+\rho]$, and the interval contains $\tau_B$.
Proposition 3 spends $\alpha/2$ on each component and adds the two half-widths. Proposition 3b spends $\alpha$ once and adds the variances, so its half-width is $z_{1-\alpha/2}\sqrt{\widehat V_\Delta+\widehat V_C}$ against $z_{1-\alpha/4}(\sqrt{\widehat V_\Delta}+\sqrt{\widehat V_C})$. When the two standard errors are equal the ratio is $z_{1-\alpha/2}\sqrt{2}/(2z_{1-\alpha/4})\approx 0.62$, so 3b is about 38\% narrower at $\alpha=0.05$. At $\alpha=0.01$, the level used when five models are tested as one family, the ratio is about 0.65 and 3b is about 35\% narrower. It gives up the union bound's distribution-free guarantee for each component and relies on the normal approximation that the Wald interval already uses. When several models are tested as one family, $\alpha$ is divided by the number of models (a Bonferroni correction), and each of the intervals above is built at that level.
Because the false-claim probability of the rule that uses $I^{\mathrm{ind}}_\tau(\rho)$ is largest when $|b_B-b_C|=\rho$ and $\tau_B=0$, the edge of the transport assumption, a simulation of the rule's false-claim rate must place the control mean at that edge; the interior of the null (equal ordinary gaps) understates it.

\subsection{Exact placebo-label check}
The private control bank also supports a design check that requires no model assumption. Suppose a control parent has (K+1) fresh forms with fixed scored outcomes $Y_1,\ldots,Y_{K+1}$. After all responses are collected, choose one form index $J$ uniformly at random and label it the pseudo-original. Define
\begin{equation*}
D^{\mathrm{pl}}(J)=Y_J-\frac{1}{K}\sum_{k\neq J}Y_k.
\end{equation*}

\medskip\noindent\textbf{Proposition 4.}\ \itshape
Conditional on the observed outcomes,
\begin{equation*}
\mathbb E_J[D^{\mathrm{pl}}(J)\mid Y_1,\ldots,Y_{K+1}]=0.
\end{equation*}

\upshape\medskip\noindent\textit{Proof.}
Let $T=\sum_{j=1}^{K+1}Y_j$. Averaging over all possible pseudo-original labels gives
\begin{equation*}
\frac{1}{K+1}\sum_{j=1}^{K+1}
\left(Y_j-\frac{T-Y_j}{K}\right)=0.
\end{equation*}

This identity yields an exact randomization check for scoring, coding, and form-role asymmetry under random labels. It does not prove that a historically public original form is exchangeable with newly written forms, so it complements rather than replaces the transport sensitivity analysis.

\subsection{Controlled-exposure validation target}
For training seed $s$, let $\widehat\Delta_{s,c}$ be the average parent-level public-form advantage after training under condition $c$. The registered policy-level effects are
\begin{equation*}
\Theta_{\mathrm{exact}}
=\mathbb E_s[
\widehat\Delta_{s,\mathrm{exact}}-
\widehat\Delta_{s,\mathrm{related}}
]
\end{equation*}

and
\begin{equation*}
\Theta_{\mathrm{para}}
=\mathbb E_s[
\widehat\Delta_{s,\mathrm{paraphrase}}-
\widehat\Delta_{s,\mathrm{related}}
].
\end{equation*}

These are effects of training-mixture policies, not item-level direct effects. An exposed training item can improve performance on related unexposed items, so exposed-versus-unexposed item summaries are diagnostic unless a credible no-interference assumption is added.

\subsection{Scope of the mathematics}
The finite-sample bounds concern the observable public-form advantage under the stated sampling design. The negative-control result concerns a sensitivity set under an explicit transport radius. Neither result proves that a proprietary model trained on a named benchmark. That boundary is part of the method, not a weakness to conceal.

\subsection{Variance-adaptive distribution-free interval (post hoc)}
The bound in Proposition 1 uses only the range of $D_i$. When a model is near
ceiling on a benchmark, almost every parent has $D_i=0$ and the bound is far
wider than the data warrant. The empirical Bernstein inequality replaces the
range with the sample variance while remaining distribution-free.

\medskip\noindent\textbf{Proposition 5.}\ \itshape
Let $s^2$ be the sample variance of $D_1,\ldots,D_n$ with $D_i\in[-1,1]$,
drawn independently (sampling with replacement) from the $N$-item benchmark. For any
$\alpha\in(0,1)$, with probability at least $1-\alpha$,
\begin{equation*}
|\widehat\Delta-\Delta_N|
\leq
\sqrt{\frac{2s^2\log(4/\alpha)}{n}}
+\frac{14\log(4/\alpha)}{3(n-1)}.
\end{equation*}

\upshape\medskip\noindent\textit{Proof.}
Set $Y_i=(D_i+1)/2\in[0,1]$. Theorem 4 of Maurer and Pontil (2009) is
one-sided: for any $\delta\in(0,1)$, with probability at least $1-\delta$,
\begin{equation*}
\mathbb E Y-\bar Y
\leq
\sqrt{\frac{2V_n\log(2/\delta)}{n}}
+\frac{7\log(2/\delta)}{3(n-1)},
\end{equation*}

where $V_n$ is the sample variance of the $Y_i$. Applying the same theorem to
$1-Y_i$, which also lies in $[0,1]$ and has the same sample variance, bounds
the opposite tail. Taking $\delta=\alpha/2$ in each direction and applying the
union bound gives a two-sided statement at level $\alpha$ in which
$\log(2/\delta)$ becomes $\log(4/\alpha)$. Since $D_i=2Y_i-1$ we have
$\widehat\Delta-\Delta_N=2(\bar Y-\mathbb E Y)$ and $V_n=s^2/4$. Substituting
and multiplying by two gives the stated bound. This proves the statement for independent draws. The audit samples without replacement, and for that design the statement is
not proved here: the convex-order comparison used for Proposition 1 applies to the
sum of the sampled values and not to the sample-variance term. The bound is supported for
sampling without replacement by the simulation below. An empirical Bernstein-Serfling
inequality that is proved for sampling without replacement is Theorem 4.3 of Bardenet and
Maillard (2015). Its constants differ from those above, and the registered branch does not use it.
Writing the two-sided statement with $\log(2/\alpha)$ would be wrong, because
the cited one-sided theorem does not supply it. The constant $\log(4/\alpha)$
widens the bound by about 9\% in the variance term and 19\% in the additive term
relative to $\log(2/\alpha)$, and every number reported for this interval in the
paper uses it.

\subsection{Use}
This proposition was added after the real-benchmark audit was analyzed, and it
is reported as a separate interval rather than substituted for the registered
one. Selecting the narrower of two intervals that each hold at level
$1-\alpha$ does not itself hold at level $1-\alpha$, so the two are always
reported side by side.
Its coverage was simulated to the same standard as every other interval in the
paper, on the six bounded outcome shapes and the finite-population sampling
design of the extended stress grid, with 4,000 repeated audits in each of 36
settings. Empirical coverage ranges from 0.9998 to 1.0000 and never falls below
the nominal level. The bound is not uniformly narrower than Proposition 1: it is
narrower in 23 of the 36 settings, with width ratios from 0.40 to 2.17, because
the additive term $14\log(4/\alpha)/(3(n-1))$ dominates when $n$ is small. At
$n=200$ it is 23\% narrower on a low-variance population and 62\% wider on a
two-point population splitting its mass between $-1$ and $+1$.
The practical consequence is specific rather than general. The registered
guardrail routes to a distribution-free bound exactly when fewer than 30 of the
sampled parents have a nonzero gap, that is, in the concentrated regime where
Proposition 5 is narrower; the regime where it loses is one the guardrail never
selects. Future audits should therefore register Proposition 5 as the
distribution-free interval on the guardrail branch and retain Proposition 1
elsewhere.

\subsection{Exact randomization test for a public-form advantage}
Proposition 5 widens what a distribution-free interval can say, but it does not
change the fact that an interval on the mean of $D_i$ uses only the range of
$D_i$ and ignores that most parents contribute exactly zero. Section 7.6 shows
the cost: below roughly two hundred audited parents the interval has no power at
any plausible effect size. For detection, as opposed to estimation, the audit
can do much better, and the ingredient is already in Proposition 4.

\subsection{Setup}
Parent $i$ has (K+1) scored forms $Y_{i0},\ldots,Y_{iK}\in\{0,1\}$, where index
0 is the public form. Let $J_i$ be the index of the form playing the original
role and define
\begin{equation*}
D_i(J_i)=Y_{iJ_i}-\frac{1}{K}\sum_{k\neq J_i}Y_{ik},
\qquad
\bar D(J)=\frac1n\sum_{i=1}^n D_i(J_i).
\end{equation*}

The observed statistic is $\bar D(0)$, the audit's estimate.

\subsection{Null hypothesis}
$H_0$: within each parent, the public form is exchangeable with that parent's
fresh forms. This is the sharp null that the published wording confers no
advantage of any kind, whether from prior exposure or from ordinary form
mismatch. It is a detection null, not an identification statement; separating
exposure from form mismatch remains the job of the transport analysis.

\medskip\noindent\textbf{Proposition 6.}\ \itshape
Draw $J_1,\ldots,J_n$ independently and uniformly from $\{0,\ldots,K\}$ and let
\begin{equation*}
p=\Pr_J\!\left(|\bar D(J)|\geq|\bar D(0)|\ \middle|\ Y\right).
\end{equation*}

Suppose $H_0$ holds in the following sense: the joint distribution of the
correctness vectors is unchanged when the forms within each parent are
relabeled, with an independent relabeling in each parent. Then
$\Pr(p\leq\alpha)\leq\alpha$ for every $\alpha\in(0,1)$ and every sample size
$n$, whatever the joint distribution otherwise is.

\upshape\medskip\noindent\textit{Proof.}
Condition on the multiset of outcomes within each parent. Under $H_0$ the
observed labeling of forms within parent $i$ is, conditionally, a uniform draw
from the labelings of that multiset, independently across parents, so the
observed $J_i=0$ has the same conditional law as a uniform draw from
$\{0,\ldots,K\}$. The observed statistic $\bar D(0)$ is therefore a draw from
the same conditional distribution that defines the reference set, and $p$ is
the upper-tail probability of a draw under its own distribution. Such a
p-value satisfies $\Pr(p\leq\alpha)\leq\alpha$, which is the standard validity
result for randomization tests (Lehmann and Romano, 2005). The
reference set is finite, so the statement is exact and not asymptotic. With
Monte Carlo evaluation over $M$ draws, using
$(1+\#\{|\bar D(J^{(m)})|\geq|\bar D(0)|\})/(1+M)$ preserves validity.
Proposition 4 is the first moment of the same reference distribution: it states
that $\mathbb E_J[D_i(J_i)]=0$, which is why the reference distribution is
centered at zero.

\subsection{What this buys and what it does not}
The test conditions on the observed correctness vectors, so parents on which
every form was answered identically contribute no variability and do not dilute
the signal, whereas they inflate the range-based bound. Appendix B.2 measures
the consequence: at one hundred audited parents the interval detects nothing at
any effect size up to ten points, while the exact test detects a ten-point
advantage in 77\% of audits on GSM8K and 95\% on ARC, at a simulated type-I error
between 0.029 and 0.052.
The test reports whether an advantage is present, not how large it is. The
intervals of Section 3.3 and Proposition 5 remain the way to bound magnitude,
and the sensitivity interval of Proposition 3 remains the way to ask how much of
any advantage could be exposure. An audit should report the test for detection
and the intervals for magnitude; neither replaces the other.

\section{Further experiments}
\subsection{A variance-adaptive bound for near-ceiling models}
When a model is near ceiling on a benchmark, very few parents show any gap. Qwen2.5-7B under Amendment 1 has 16 nonzero gaps out of 200. The guardrail then makes the Hoeffding bound primary, and at n = 200 that bound is $\pm$ 0.192 whatever the data look like, because it uses only the range of $D_i$. A variance-adaptive bound keeps the distribution-free guarantee. For $D_i$ in [-1, 1], the empirical Bernstein inequality (Maurer and Pontil, 2009) applied to the rescaled variable gives, with probability at least 1 - $\alpha$,

\begin{equation*}
|\widehat\Delta-\Delta_N|
\leq
\sqrt{\frac{2s^2\log(4/\alpha)}{n}}
+\frac{14\log(4/\alpha)}{3(n-1)},
\end{equation*}

where $s^2$ is the sample variance of the parent gaps. The log term is log(4/$\alpha$) and not log(2/$\alpha$) because the Maurer-Pontil result is one-sided, so covering both tails costs $\alpha$/2 in each direction. The bound is proved for independent draws. For sampling without replacement it is checked by simulation and not proved here: the comparison used for Proposition 1 covers the sum of the sampled values and not the sample-variance term. An empirical Bernstein-Serfling inequality that is proved for sampling without replacement is Theorem 4.3 of Bardenet and Maillard (2015). Its constants differ from those above, and the registered branch does not use it. The derivation is Proposition 5 in Appendix A. The bound was added after the real audit was analyzed. It is reported beside the registered interval and never substituted for it, because choosing the smaller of two intervals after seeing both does not keep the nominal level.

We held it to the standard of Section 6 and simulated its coverage on the same six outcome shapes, two benchmark sizes and three audit fractions, with 4,000 audits per setting (Table 15). The populations in this run are drawn afresh from the same shape distributions, with a different seed, so its Wald coverage range (0.9390 to 0.9563) differs from the range in Section 6.3 (0.9145 to 0.9575); Wald coverage at small n depends strongly on how many nonzero gaps a realized population happens to contain.

\begin{table}[t]
\centering
\small
\caption{Coverage of the three intervals across the 36 stress settings (4,000 audits each), and the width of the Bernstein bound relative to the Hoeffding bound at the audit's own sample size, n = 200 in a 1,000-item benchmark.}
\label{tab:15}
\resizebox{\ifdim\width>\textwidth \textwidth\else \width\fi}{!}{%
\begin{tabular}{llrl}
\toprule
\textbf{Interval} & \textbf{Coverage range} & \textbf{Settings below 0.95} & \textbf{Width vs Hoeffding at n = 200} \\
\midrule
Hoeffding & 0.9945 to 1.0000 & 0 & 1.00 by definition \\
Empirical Bernstein & 0.9998 to 1.0000 & 0 & 0.77 to 1.62, depending on the shape \\
Finite-population Wald & 0.9390 to 0.9563 & 24 & not distribution-free \\
\bottomrule
\end{tabular}}
\end{table}

The Bernstein bound did not under-cover in any of the 36 settings: its minimum coverage is 0.9998, and like the Hoeffding bound it is conservative. It is not uniformly narrower. It is narrower than Hoeffding in 23 of the 36 settings, with width ratios from 0.40 to 2.17. The additive term 14 log(4/$\alpha$) / (3(n-1)) dominates at small audit sizes, so at n = 50 the Bernstein bound is wider in every shape. At n = 200 it is 23\% narrower on the low-variance shape, 12\% narrower on the sparse and near-null shapes, level on the right-skewed shape, 12\% wider on the balanced shape and 62\% wider on the high-variance two-point shape. At n = 100 the additive term alone is 0.207, against a Hoeffding half-width of 0.272, so the Bernstein bound is narrower only if the sample variance is below about 0.048.

This pattern lines up with the guardrail. The guardrail fires when fewer than 30 of 200 sampled parents have a nonzero gap, the concentrated low-variance regime in which the Bernstein bound is narrower. On the audited data it is narrower than Hoeffding for eight of ten pairs, with half-widths from 0.148 to 0.197 against 0.192. The two exceptions are the two GSM8K pairs with the most nonzero gaps, SmolLM2-1.7B and Qwen2.5-1.5B, and both are Wald-primary, so the guardrail would never route to the Bernstein bound there. On both pairs where the guardrail fires it is narrower. In the Amendment 1 run for Qwen2.5-7B, the near-ceiling case, it gives 0.144 against 0.192, a 25\% reduction. We recommend that future audits register the empirical Bernstein bound as the distribution-free interval on the guardrail branch and keep Hoeffding elsewhere. This is a property of these data and of the low-variance regime, not a general ordering.

\subsection{An exact randomization test for small audits}
The interval on the mean gap has no power below about 200 parents, because a bounded-difference bound uses only the range of $D_i$, so parents answered identically on every form widen the bound exactly as much as informative ones. For detection, the audit can instead condition on the observed correctness vectors. Proposition 4 says that if the pseudo-original role is assigned at random among a parent's forms the expected gap is exactly zero. Proposition 6 uses the whole reference distribution. Under the sharp null that the public form is exchangeable with its fresh forms within each parent, relabeling which form plays the original role, independently in each parent, generates the exact null distribution of the mean gap, and the p-value is valid at any sample size.

\begin{table}[t]
\centering
\small
\caption{Detection rate of the exact randomization test against the interval of Section 7.6, same planted advantages, same resampled banks, 2,000 audits per cell. The test's simulated type-I error across all sizes and both benchmarks runs from 0.029 to 0.052 against a nominal 0.05.}
\label{tab:16}
\resizebox{\ifdim\width>\textwidth \textwidth\else \width\fi}{!}{%
\begin{tabular}{lrrrr}
\toprule
\textbf{Benchmark} & \textbf{Audited items} & \textbf{delta} & \textbf{Interval} & \textbf{Exact test} \\
\midrule
GSM8K & 50 & 0.10 & 0.00 & 0.43 \\
GSM8K & 50 & 0.20 & 0.00 & 0.92 \\
GSM8K & 100 & 0.05 & 0.00 & 0.32 \\
GSM8K & 100 & 0.10 & 0.00 & 0.77 \\
GSM8K & 200 & 0.05 & 0.56 & 0.61 \\
GSM8K & 200 & 0.10 & 0.95 & 0.97 \\
ARC & 50 & 0.10 & 0.00 & 0.64 \\
ARC & 100 & 0.05 & 0.00 & 0.52 \\
ARC & 100 & 0.10 & 0.00 & 0.95 \\
ARC & 200 & 0.05 & 0.84 & 0.86 \\
\bottomrule
\end{tabular}}
\end{table}

Table 16 compares the test with the interval at the sizes where they differ. Simulated type-I error stays between 0.029 and 0.052 at every size on both benchmarks, against a nominal 0.05. At 200 parents and above the two agree closely. At 100 parents a ten-point advantage goes from undetectable to 77\% detection on GSM8K and 95\% on ARC. Applied to the audited data the test rejects for none of the ten pairs, with p-values from 0.16 to 1.00, so the nulls survive a procedure with far more power than the interval.

What the test is exact for needs care, because the sharp null is a joint one. It says a parent's forms are exchangeable for this model, which is true only when there is neither contamination nor ordinary form mismatch. The control banks show that mismatch is small but not zero, with form gaps from -0.088 to +0.033 across the ten pairs. The type-I simulation cannot see this, because it relabels the real correctness vectors and thereby imposes exchangeability. Table 17 measures the cost: audits of 200 parents with no contamination at all and a fresh-form penalty of a given size.

\begin{table}[t]
\centering
\small
\caption{What ordinary form mismatch does to the exact test. No contamination is present in any row; every rejection is caused by the fresh forms being slightly harder than the public one. The largest form gap is the largest measured on any control bank in this audit.}
\label{tab:17}
\resizebox{\ifdim\width>\textwidth \textwidth\else \width\fi}{!}{%
\begin{tabular}{rr}
\toprule
\textbf{Form gap in the data} & \textbf{Rejection rate at the 0.05 level} \\
\midrule
0.000 (exchangeable) & 0.039 \\
0.020 & 0.107 \\
0.050 & 0.315 \\
0.088 (largest measured) & 0.689 \\
\bottomrule
\end{tabular}}
\end{table}

The first row is the exactness the paper claims. The last row is the same test, with nothing to detect, rejecting on more than two thirds of audits. A rejection therefore cannot be read as evidence of a public-form advantage. The test can be used in one direction only: a non-rejection fails to reject the sharp null of no public-form advantage of any kind, contamination or form mismatch, exactly and at any sample size. It does not bound the size of an advantage. The registered analysis of this paper is unchanged, and the test is reported beside it.

\subsection{Rewrite quality}
The audit rests on the fresh forms being faithful, and the authors' reading came after the results. We therefore ask what the results say about the items, independently of anyone's reading. A rewrite that is harder than its original, or that drops needed information, should trip more than one model. For each parent we counted the models that answered the public form correctly and both fresh forms incorrectly, and tested the count of parents with at least two such models against a null that permutes each model's failures across parents. We ran the mirror-image test too, for parents where the public form was wrong and both rewrites right, which is what a rewrite that leaked a hint would look like. Table 18 gives both tests.

\begin{table}[t]
\centering
\small
\caption{Parents on which at least two of the five models show the pattern, against a permutation null with 20,000 draws.}
\label{tab:18}
\resizebox{\ifdim\width>\textwidth \textwidth\else \width\fi}{!}{%
\begin{tabular}{llrrr}
\toprule
\textbf{Benchmark} & \textbf{Pattern} & \textbf{Observed} & \textbf{Null mean} & \textbf{p} \\
\midrule
GSM8K & rewrites harder & 4 & 3.37 & 0.45 \\
GSM8K & rewrites easier & 2 & 3.29 & 0.88 \\
ARC & rewrites harder & 3 & 0.70 & 0.02 \\
ARC & rewrites easier & 0 & 0.30 & 1.00 \\
\bottomrule
\end{tabular}}
\end{table}

On GSM8K, failures do not cluster: four parents show the pattern where chance predicts 3.4, and the mirror test is below chance. On ARC, three parents cluster where chance predicts 0.7. All three rewrites preserve the information, the options and the key, and none adds a hint. What the three share is a short or weakly cued stem, including one whose original is the seven-word fragment “The suspension system of a truck includes the”. Under length-normalized option scoring, a stem that carries little information leaves the choice largely to the option text, so small wording changes can move the choice. One of the three also contains a defect in the item: the second rewrite of that parent begins “Included in the suspension system of a truck is the”, where every option is a compound plural, so the completion is ungrammatical whichever option is chosen. The defect affects all four options about equally, but it makes that fresh form noisier than its original for reasons unrelated to contamination. We left the item as it is, and the three parents are in the released review packet.

A second failure mode shows an asymmetry in the registered verifier. ARC fresh stems must have Jaccard overlap below 0.80 with the original, and twelve first-pass rewrites were rejected under that rule. GSM8K forms were checked only for exact identity, so the rule was never applied to them. Applying it afterward, 21 of the 200 GSM8K parents have at least one fresh form at or above 0.80, the closest reaching 0.895. On ARC no parent reaches 0.80 and the closest is 0.789. A near-copy should be answered like its original whether or not the model has seen either, which pulls the parent's gap toward zero. Table 19 removes those parents and recomputes.

\begin{table}[t]
\centering
\small
\caption{Sensitivity of the GSM8K audit to rewrite depth. The third column drops the 21 parents whose closest fresh form meets or exceeds the ARC dissimilarity threshold. The last column is the correlation, across all 200 parents, between a parent's rewrite similarity and its measured gap.}
\label{tab:19}
\resizebox{\ifdim\width>\textwidth \textwidth\else \width\fi}{!}{%
\begin{tabular}{lrrrrr}
\toprule
\textbf{Model} & \textbf{$\hat{\Delta}$, all 200} & \textbf{$\hat{\Delta}$, deep rewrites only} & \textbf{Shift} & \textbf{r (similarity, gap)} & \textbf{p} \\
\midrule
OLMo-2-7B & +0.015 & +0.028 & +0.013 & +0.029 & 0.69 \\
Phi-3-mini & +0.018 & +0.017 & -0.001 & +0.023 & 0.75 \\
Qwen2.5-1.5B & -0.025 & -0.020 & +0.005 & -0.035 & 0.62 \\
Qwen2.5-7B & -0.005 & +0.003 & +0.008 & -0.072 & 0.31 \\
SmolLM2-1.7B & +0.045 & +0.050 & +0.005 & -0.004 & 0.96 \\
\bottomrule
\end{tabular}}
\end{table}

The largest shift is 0.013, against half-widths of 0.035 to 0.058, so no conclusion in Table 4 depends on the shallow rewrites. The correlation between rewrite similarity and gap is within noise for every model, pooled r = -0.014 on GSM8K and -0.039 on ARC. The 21 near-copy parents show a mean gap of -0.043 against +0.016 for the rest, the opposite of the direction the concern predicts and not significant (Welch p = 0.18). Every one of the 21 is the parent's first fresh form, never the second. The median near-copy changes 11\% of its words, and the most extreme changes one word in nineteen, turning `\texttt{A bag has a 5\% discount'' into }`A jacket has a 5\% discount''. The first drafting pass was a light substitution of names and objects and the second restructured the sentence. We report this as a defect in the verifier and the drafting procedure, not in the bank, and we did not modify the bank. The verifier now computes the GSM8K similarity and lists the 21 items. An audit built from this code should apply the threshold to both benchmarks before freezing, and should check that a parent's two rewrites are of comparable depth.

\subsection{Budgeted audits}
How small can an audit be? Using the 200 audited parents as a finite population, we drew 2,000 subsamples without replacement at each budget and recorded the finite-population Wald half-width, whether the subsample reproduced the full-sample ordering of all five models by fresh-form score, and the share of the ten model pairs it ordered correctly (Table 20).

\begin{table}[t]
\centering
\small
\caption{Budgeted-audit curves (mean over models and 2,000 subsamples). `\texttt{Exact order'' is recovery of the full five-model ordering; }`pairs correct'' is the share of the ten model pairs ordered as the full sample orders them.}
\label{tab:20}
\resizebox{\ifdim\width>\textwidth \textwidth\else \width\fi}{!}{%
\begin{tabular}{lrrrr}
\toprule
\textbf{Benchmark} & \textbf{Audited items} & \textbf{Mean half-width} & \textbf{Exact order} & \textbf{Pairs correct} \\
\midrule
GSM8K & 25 & 0.142 & 0.39 & 0.92 \\
GSM8K & 50 & 0.101 & 0.50 & 0.95 \\
GSM8K & 100 & 0.070 & 0.62 & 0.97 \\
GSM8K & 150 & 0.056 & 0.73 & 0.98 \\
ARC & 25 & 0.093 & 0.15 & 0.78 \\
ARC & 50 & 0.067 & 0.25 & 0.86 \\
ARC & 100 & 0.047 & 0.49 & 0.94 \\
ARC & 150 & 0.038 & 0.67 & 0.97 \\
\bottomrule
\end{tabular}}
\end{table}

Pairwise ordering is cheap: 100 audited GSM8K items order 97\% of model pairs as the full audit does, and 100 ARC items order 94\%. Recovering the exact ordering of all five models is much harder, 0.62 and 0.49 at the same budget, and it stays below one until the audit is complete. The reason is in the scores. On GSM8K, OLMo-2-7B and Qwen2.5-7B differ by 0.010 in fresh-form score, and on ARC, Phi-3-mini and OLMo-2-7B differ by 0.010. Both gaps are smaller than the audit's half-width at every budget below the full sample, so no audit of that size can order those pairs, and one swapped adjacent pair destroys the exact ordering. A few dozen items suffice to separate models that differ by more than the half-width in that row, and no audit separates models that differ by less than its own half-width. Reporting the audit-adjusted score with its interval, not a bare rank, makes that limit visible.

\subsection{The failed ARC study and the corrected leak}
Table 21 gives the failed study at full length. The registered manipulation check is the difference in differences on the trained wording, required to be at least +0.05, and it came in at -0.220. The deep-paraphrase row is reported for completeness and contributes nothing once the check has failed. Table 22 gives the corrected leak, where only the exposure record changed. In that table the deep-paraphrase arm used the control group's change on fresh forms as its counterfactual, which is the substitution that Section 8.3 replaces with a direct comparison.

\begin{table}[t]
\centering
\small
\caption{The failed manipulation. Accuracy on the ARC items that were leaked, beside the ones that were not, before and after fine-tuning. The difference in differences on the trained wording is the manipulation check, and the registration required it to be at least +0.05.}
\label{tab:21}
\resizebox{\ifdim\width>\textwidth \textwidth\else \width\fi}{!}{%
\begin{tabular}{llrrrl}
\toprule
\textbf{Parents} & \textbf{Form} & \textbf{Before} & \textbf{After} & \textbf{Difference in differences} & \textbf{95\% CI} \\
\midrule
Exposed (n = 100) & public, trained on & 0.460 & 0.280 & \textbf{-0.220} & [-0.350, -0.090] \\
Exposed (n = 100) & fresh, never trained on & 0.420 & 0.375 & -0.090 & [-0.205, +0.025] \\
Exposed (n = 93) & deep paraphrase & 0.387 & 0.301 & -0.131 & [-0.231, -0.032] \\
Unexposed (n = 100) & public & 0.400 & 0.440 & — & — \\
Unexposed (n = 100) & fresh & 0.400 & 0.445 & — & — \\
\bottomrule
\end{tabular}}
\end{table}

\begin{table}[t]
\centering
\small
\caption{The corrected leak. Accuracy on the ARC items that were leaked, beside the hundred that were not, before and after fine-tuning. The rightmost column is the difference in differences against the untouched parents.}
\label{tab:22}
\resizebox{\ifdim\width>\textwidth \textwidth\else \width\fi}{!}{%
\begin{tabular}{llrrrl}
\toprule
\textbf{Parents} & \textbf{Form} & \textbf{Before} & \textbf{After} & \textbf{Difference in differences} & \textbf{95\% CI} \\
\midrule
Exposed (n = 100) & public, trained on & 0.460 & 1.000 & \textbf{+0.490} & [+0.360, +0.620] \\
Exposed (n = 100) & fresh, never trained on & 0.420 & 0.980 & \textbf{+0.510} & [+0.395, +0.625] \\
Exposed (n = 93) & deep paraphrase & 0.376 & 0.602 & +0.176 & [+0.039, +0.315] \\
Unexposed (n = 100) & public & 0.400 & 0.450 & — & — \\
Unexposed (n = 100) & fresh & 0.400 & 0.450 & — & — \\
\bottomrule
\end{tabular}}
\end{table}

\subsection{The first GSM8K positive control in detail}
Table 23 gives the four accuracies before and after fine-tuning for the first GSM8K study, and Table 24 the differences in differences. The other never-contaminated sets behaved like the unexposed hundred. The ARC audit, on a benchmark the contamination never touched, moved +0.023. The private GSM8K control bank moved +0.067 and the ARC control bank +0.013. None differs significantly from zero (p between 0.27 and 1.00), and the exact 0.000 on the unexposed half is partly luck, so the supported claim is that nine points of non-specific damage to the model produced no detectable movement in the paired difference anywhere it was not contaminated.

\begin{table}[t]
\centering
\small
\caption{Where the contamination effect went. Accuracy before and after exposure, on the hundred leaked parents and the hundred untouched ones. The measurable gap is the difference between the public and fresh columns, which is what any paraphrase audit observes.}
\label{tab:23}
\resizebox{\ifdim\width>\textwidth \textwidth\else \width\fi}{!}{%
\begin{tabular}{llrrr}
\toprule
\textbf{Parents} & \textbf{Form} & \textbf{Before} & \textbf{After} & \textbf{Change} \\
\midrule
Exposed (n = 100) & public, trained on & 0.460 & 0.610 & \textbf{+0.150} \\
Exposed (n = 100) & fresh, never trained on & 0.485 & 0.570 & \textbf{+0.085} \\
Exposed (n = 100) & \emph{measurable gap} & -0.025 & +0.040 & \textbf{+0.065} \\
Unexposed (n = 100) & public & 0.490 & 0.400 & -0.090 \\
Unexposed (n = 100) & fresh & 0.515 & 0.425 & -0.090 \\
Unexposed (n = 100) & \emph{measurable gap} & -0.025 & -0.025 & \textbf{+0.000} \\
\bottomrule
\end{tabular}}
\end{table}

\begin{table}[t]
\centering
\small
\caption{Difference-in-differences estimates for the effect of leaking an item's public wording, with bootstrap intervals over parents.}
\label{tab:24}
\resizebox{\ifdim\width>\textwidth \textwidth\else \width\fi}{!}{%
\begin{tabular}{lrrrl}
\toprule
\textbf{Quantity} & \textbf{Exposed} & \textbf{Unexposed} & \textbf{Difference in differences} & \textbf{95\% CI} \\
\midrule
Change in public-form accuracy & +0.150 & -0.090 & \textbf{+0.240} & [+0.080, +0.400] \\
Change in fresh-form accuracy & +0.085 & -0.090 & \textbf{+0.175} & [+0.035, +0.315] \\
Change in the measurable gap & +0.065 & +0.000 & \textbf{+0.065} & — \\
\bottomrule
\end{tabular}}
\end{table}

We also tried and rejected a selection-based analysis, because it is an easy mistake. Conditioning on the items the model newly answered correctly on the leaked wording gives an apparently sharp estimate: for those 25 items, fresh-form accuracy rose by +0.400 with a 95\% interval of [+0.242, +0.558]. The identical calculation on the unexposed half, where no leak occurred, gives +0.250 on twelve items. Most of the apparent effect is selection, since conditioning on an item that improved on one form picks items that improved on correlated forms. The difference between the two, +0.150, is not significant (p = 0.28). The difference-in-differences estimator avoids this because it conditions on assignment, not on outcome.

\subsection{Study 9: four GSM8K positive controls under four training seeds}
Study 9 (\texttt{POSITIVE\_\allowbreak CONTROL\_\allowbreak STUDY9\_\allowbreak REGISTRATION.md}) repeated the four GSM8K positive controls of Section 8.1 under three further training seeds each, 20261024, 20261025 and 20261026, with one session per model and seed. The leaked items, the mixture and the audited parents were the same in every run, and only the training seed changed. The training script gained a seed argument; when it is absent the training mixture is byte for byte the one used before, which we confirmed from its SHA-256. Each of the twelve new runs trained to a loss within a few percent of the registered run, produced a complete 600-form audit and was analyzed. None failed or was rerun. The first two kernels (Qwen2.5-1.5B and StableLM-2-1.6B, seed 20261024) were launched before the estimator and the decision wording of the registration were finalized. Those parts were finalized before any Study 9 result was read, after a review pointed out that an earlier version of the pooling resampled the shared parents independently across models. Table 25 lists all sixteen runs.

\begin{table}[t]
\centering
\small
\caption{Every run of Study 9. The registered run is the one of Studies 1 to 3 (training seed 20260822); the other three per model were added by Study 9. Differences in differences against the unexposed control parents of the same run.}
\label{tab:25}
\resizebox{\ifdim\width>\textwidth \textwidth\else \width\fi}{!}{%
\begin{tabular}{llrrr}
\toprule
\textbf{Model} & \textbf{Training seed} & \textbf{DiD public} & \textbf{DiD fresh} & \textbf{Visible gap} \\
\midrule
Qwen2.5-1.5B & 20260822 (registered) & +0.240 & +0.175 & +0.065 \\
Qwen2.5-1.5B & 20261024 & +0.210 & +0.055 & +0.155 \\
Qwen2.5-1.5B & 20261025 & +0.230 & +0.170 & +0.060 \\
Qwen2.5-1.5B & 20261026 & +0.290 & +0.215 & +0.075 \\
SmolLM2-1.7B & 20260822 (registered) & +0.100 & +0.040 & +0.060 \\
SmolLM2-1.7B & 20261024 & +0.110 & +0.035 & +0.075 \\
SmolLM2-1.7B & 20261025 & +0.180 & +0.040 & +0.140 \\
SmolLM2-1.7B & 20261026 & +0.110 & +0.045 & +0.065 \\
OLMo-2-1B & 20260822 (registered) & +0.500 & +0.350 & +0.150 \\
OLMo-2-1B & 20261024 & +0.440 & +0.435 & +0.005 \\
OLMo-2-1B & 20261025 & +0.420 & +0.320 & +0.100 \\
OLMo-2-1B & 20261026 & +0.480 & +0.340 & +0.140 \\
StableLM-2-1.6B & 20260822 (registered) & +0.200 & +0.230 & -0.030 \\
StableLM-2-1.6B & 20261024 & +0.120 & +0.170 & -0.050 \\
StableLM-2-1.6B & 20261025 & +0.210 & +0.150 & +0.060 \\
StableLM-2-1.6B & 20261026 & +0.190 & +0.300 & -0.110 \\
\bottomrule
\end{tabular}}
\end{table}

The pooled quantities are means over models of each model's mean over its runs, and the invisible share is the pooled fresh-form effect divided by the pooled public-form effect. The registered interval resamples the exposed and the unexposed parents once per draw and applies that draw to every model and seed, because the same parents were scored in every run. It then adds a seed component for each model: a t variable with three degrees of freedom times the model's between-run standard deviation divided by two, which is the sampling error of the mean of four runs. The public and fresh components are drawn independently, which is the conservative choice for the ratio. There are 20,000 draws. Three comparison intervals show where the width comes from. The estimator of Study 3, which resamples parents independently for each model and uses the registered runs only, gives [52\%, 110\%] for the share. Shared parents on the registered runs only gives [54\%, 108\%]. Correlating the two seed components gives [54\%, 105\%]. The registered interval is [50\%, 113\%], so the seed component widens it a little on both sides and the three extra seeds do not narrow it.

The registered consequences were applied as written. The pooled interval for the effect on unseen paraphrases, [+0.125, +0.260], excludes zero, so transfer is established under seed variation. The lower limit of the share, 49.6\%, is below one half, so the paper says that a substantial share of the effect is invisible and does not say that most of it is. The point estimate moved by one point of share, below the five-point threshold for reporting the earlier figure beside the new one. One line of the pooling script, which excludes draws with a pooled public-form effect at or below 0.05 from the ratio, is not in the registration. It removes one of 20,000 draws and changes no reported number.

\subsection{The registered larger audit, not yet run}
The registration is \texttt{REAL\_\allowbreak AUDIT\_\allowbreak EXTENSION\_\allowbreak REGISTRATION.md}, with four dated amendments and the analysis code fixed by hash. It takes 800 GSM8K parents, the first 200 and 600 more drawn without replacement, against a control bank of 900 never-public problems, with the same five models and decoding at a 768-token budget. We describe the design because the simulations behind it are results of this paper.

Two design findings came from checking the registration, not from data. First, the claim rule should be Proposition 3b with a Bonferroni correction across the models. It is one-sided: a model is reported as showing an exposure-consistent advantage only if the public-form advantage is positive, its primary interval excludes zero, and the lower end of the sensitivity interval at the radius of 0.02 lies above zero. At the edge of the transport assumption, where a false claim is most likely, the unadjusted rule claims in 1.1\% to 1.7\% of audits per model, and that is too much across five models. Second, the control bank must be large. With 300 control items three models would have had almost no power at ten points (Phi-3-mini 0.1\%, Qwen2.5-1.5B 0.5\% and Qwen2.5-7B 0.0\%): their control gaps are nonzero on few items, so the guardrail of Section 6.3 binds on the control bank and its interval stays wide however large the audit is. With 900 items every model has power of about 95\% or more at ten points, the lowest being SmolLM2-1.7B at 95.0\% (Table 26), and the false-claim rate under Bonferroni is 0.3\% per model at the edge. Where the guardrail still binds, the empirical Bernstein interval carries the rule.

\begin{table}[t]
\centering
\small
\caption{Simulated detection rate of the registered claim rule of the extension (Proposition 3b at the Bonferroni level, 800 audited parents), by planted public-form advantage, with the control bank at 900 items. Resampled from the first audit's gap distributions, with the two Qwen models' gaps taken from its 768-token reruns (the budget of the extension) and the other three models' from its 320-token run; 4,000 simulated audits per cell. The last column shows the same rule with a 300-item control bank, the design that was dropped.}
\label{tab:26}
\resizebox{\ifdim\width>\textwidth \textwidth\else \width\fi}{!}{%
\begin{tabular}{lrrrr}
\toprule
\textbf{Case} & \textbf{5 points} & \textbf{10 points} & \textbf{15 points} & \textbf{10 points, 300 control items} \\
\midrule
OLMo-2-7B & 31.4\% & 99.8\% & 100.0\% & 92.2\% \\
SmolLM2-1.7B & 14.5\% & 95.0\% & 100.0\% & 56.5\% \\
Phi-3-mini & 66.8\% & 99.9\% & 100.0\% & 0.1\% \\
Qwen2.5-1.5B & 9.0\% & 97.7\% & 100.0\% & 0.5\% \\
Qwen2.5-7B & 0.0\% & 100.0\% & 100.0\% & 0.0\% \\
All five pooled & 36.8\% & 100.0\% & 100.0\% & 92.3\% \\
Pooled, benchmark's ordinary gap 0.02 below the control's & 2.6\% & 96.8\% & 100.0\% & 69.0\% \\
\bottomrule
\end{tabular}}
\end{table}

These are simulations from the first audit's gap distributions, not forecasts. The pilot data already show one case the registration had to anticipate. In the first audit's 768-token rerun, Qwen2.5-1.5B has a public-form advantage of -0.040 on 200 parents and a control gap of +0.058 on 60 items, so their difference is -0.098, with a 95\% interval of [-0.177, -0.020]. That lies below minus the radius. The registration fixes the reading in advance: such a model is reported as a failure of the transport assumption in the conservative direction, not as evidence about exposure, and its control gap is not used as a proxy.

\subsection{Status of every analysis}
Table 27 lists each analysis of the paper by what was fixed before its results were seen. Registered means the protocol and the decision rule were written down before the run. A deviation from a registration is described where the analysis is reported.

\begin{table}[t]
\centering
\small
\caption{Registration status of each analysis in this paper.}
\label{tab:27}
\resizebox{\ifdim\width>\textwidth \textwidth\else \width\fi}{!}{%
\begin{tabular}{lll}
\toprule
\textbf{Analysis} & \textbf{Sections} & \textbf{Status} \\
\midrule
Controlled-exposure validation in two model families & 5 & Registered before the runs \\
Design simulations of interval coverage and transport-radius failure & 6 & Simulation results, not tied to the audited data \\
Audit of five models: estimates, intervals, guardrail, sensitivity interval, claim rule (320-token budget) & 7.2, 7.4 & Registered before any audited model saw an audit item \\
Rerun of the two Qwen models at 768 tokens with a parser that accepts boxed answers & 7.3 & Amendment 1, declared after the primary results and before the rerun \\
Baseline comparisons on the same forms & 7.5 & Registered before any audited model saw an audit item; the registration fixed no multiplicity correction \\
Power calibration and claim-rule detection rates resampled from the audited gaps; item-quality tests and rewrite-depth sensitivity; exact randomization test; budgeted audits; three readings of the control bank & 7.4, 7.6, B.2 to B.4 & Not in the registration; reported beside the registered analysis, which is unchanged \\
Empirical Bernstein interval for sparse audits & 7.2, B.1 & Post hoc, added after the primary analysis; reported beside the registered interval and never in its place \\
Truncation check on the three models that Amendment 1 did not rerun & 7.3 & Post hoc, not registered \\
Rerun of those three models at 768 tokens (Arm A), and ARC scored from generated letters for all five (Arm B) & 7.3 & Amendment 2, declared before any model was queried on either arm \\
Positive controls on GSM8K: Studies 1 to 3 & 8.1 & Registered before the runs; one training run per model \\
Study 9: three further seeds for the four GSM8K models & 8.1, B.7 & Seeds and pooling registered; the estimator wording was finalized after the first two runs were launched and before any result was read \\
ARC positive control (Study 4) & 8.2 & Registered; failed its manipulation check; run before the reading that its registration required \\
ARC corrected leak (Study 5) & 8.2 & Registered, and run after the stopping rule of Study 4 said no further ARC studies \\
ARC complete deep-paraphrase bank and no-leak run (Studies 6 and 7) & 8.3 & Registered before the runs \\
ARC seed replication (Study 8) & 8.3 & Registered, with the rule against extra seeds overridden by the authors; every seed reported \\
Larger audit of 800 GSM8K parents & 7.7, B.8 & Registered with four amendments before any model run; not run \\
\bottomrule
\end{tabular}}
\end{table}


\vskip 0.2in
\begin{thebibliography}{99}
\bibitem[Bardenet(2015)]{ref1} Bardenet, R., \& Maillard, O.-A. (2015). Concentration inequalities for sampling without replacement. \emph{Bernoulli, 21}(3), 1361-1385. https://doi.org/10.3150/14-BEJ605
\bibitem[Cheng(2025)]{ref2} Cheng, Y., Wang, W., Moayeri, M., \& Feizi, S. (2025). DyePack: Provably flagging test set contamination in LLMs using backdoors. \emph{Proceedings of the 2025 Conference on Empirical Methods in Natural Language Processing}, 15367-15384. https://doi.org/10.18653/v1/2025.emnlp-main.776
\bibitem[Cochran(1977)]{ref3} Cochran, W. G. (1977). \emph{Sampling techniques} (3rd ed.). Wiley.
\bibitem[Dekoninck(2024)]{ref4} Dekoninck, J., Muller, M. N., \& Vechev, M. (2024a). ConStat: Performance-based contamination detection in large language models. \emph{Advances in Neural Information Processing Systems, 37}.
\bibitem[Dekoninck(2024)]{ref5} Dekoninck, J., Muller, M. N., Baader, M., Fischer, M., \& Vechev, M. (2024b). Evading data contamination detection for language models is (too) easy. \emph{arXiv preprint} arXiv:2402.02823. https://doi.org/10.48550/arXiv.2402.02823
\bibitem[Fu(2025)]{ref6} Fu, Y., Uzuner, Ö., Yetişgen, M., \& Xia, F. (2025). Does data contamination detection work (well) for LLMs? A survey and evaluation on detection assumptions. \emph{Findings of the Association for Computational Linguistics: NAACL 2025}. https://doi.org/10.48550/arXiv.2410.18966
\bibitem[Hoeffding(1963)]{ref7} Hoeffding, W. (1963). Probability inequalities for sums of bounded random variables. \emph{Journal of the American Statistical Association, 58}(301), 13-30. https://doi.org/10.1080/01621459.1963.10500830
\bibitem[Lehmann(2005)]{ref8} Lehmann, E. L., \& Romano, J. P. (2005). \emph{Testing statistical hypotheses} (3rd ed.). Springer.
\bibitem[Lipsitch(2010)]{ref9} Lipsitch, M., Tchetgen Tchetgen, E., \& Cohen, T. (2010). Negative controls: A tool for detecting confounding and bias in observational studies. \emph{Epidemiology, 21}(3), 383-388. https://doi.org/10.1097/EDE.0b013e3181d61eeb
\bibitem[Maurer(2009)]{ref10} Maurer, A., \& Pontil, M. (2009). Empirical Bernstein bounds and sample-variance penalization. \emph{Proceedings of the 22nd Conference on Learning Theory}. https://doi.org/10.48550/arXiv.0907.3740
\bibitem[Opoku(2026)]{ref11} Opoku, J., \& Banahene, D. (2026a). PromptShift-CRC: Drift-aware conformal risk control for foundation models under prompt and domain shift. \emph{arXiv preprint} arXiv:2606.15964. https://doi.org/10.48550/arXiv.2606.15964
\bibitem[Opoku(2026)]{ref12} Opoku, J., \& Banahene, D. (2026b). ToolChain-CRC: Conformal risk control for agentic AI under retrieval and tool-use drift. \emph{arXiv preprint} arXiv:2606.18467. https://doi.org/10.48550/arXiv.2606.18467
\bibitem[Oren(2024)]{ref13} Oren, Y., Meister, N., Chatterji, N., Ladhak, F., \& Hashimoto, T. B. (2024). Proving test set contamination in black box language models. \emph{International Conference on Learning Representations}. https://doi.org/10.48550/arXiv.2310.17623
\bibitem[Serfling(1974)]{ref14} Serfling, R. J. (1974). Probability inequalities for the sum in sampling without replacement. \emph{The Annals of Statistics, 2}(1), 39-48. https://doi.org/10.1214/aos/1176342611
\bibitem[Sun(2025)]{ref15} Sun, Y., Wang, H., Li, D., Wang, G., \& Zhang, H. (2025). The emperor's new clothes in benchmarking? A rigorous examination of mitigation strategies for LLM benchmark data contamination. \emph{Proceedings of the 42nd International Conference on Machine Learning} (Proceedings of Machine Learning Research, Vol. 267, pp. 57728-57753). https://proceedings.mlr.press/v267/sun25t.html
\bibitem[Tu(2024)]{ref16} Tu, S., Zhu, K., Bai, Y., Yao, Z., Hou, L., \& Li, J. (2024). DICE: Detecting in-distribution contamination in LLM's fine-tuning phase for math reasoning. \emph{arXiv preprint} arXiv:2406.04197. https://doi.org/10.48550/arXiv.2406.04197
\bibitem[Xiao(2026)]{ref17} Xiao, X., \& Cheng, Y. (2026). Contamination inflates scores but rarely reorders large language model leaderboards. \emph{arXiv preprint} arXiv:2609.02899. https://doi.org/10.48550/arXiv.2609.02899
\bibitem[Yang(2023)]{ref18} Yang, S., Chiang, W.-L., Zheng, L., Gonzalez, J. E., \& Stoica, I. (2023). Rethinking benchmark and contamination for language models with rephrased samples. \emph{arXiv preprint} arXiv:2311.04850. https://doi.org/10.48550/arXiv.2311.04850
\bibitem[Zhang(2024)]{ref19} Zhang, H., Da, J., Lee, D., Robinson, V., Wu, C., Song, W., Zhao, T., Raja, P., Zhuang, C., Slack, D., Lyu, Q., Hendryx, S., Kaplan, R., Lunati, M., \& Yue, S. (2024). A careful examination of large language model performance on grade school arithmetic. \emph{Advances in Neural Information Processing Systems, 37}. https://doi.org/10.48550/arXiv.2405.00332
\bibitem[Zhang(2024)]{ref20} Zhang, H., Lin, Y., \& Wan, X. (2024). PaCoST: Paired confidence significance testing for benchmark contamination detection in large language models. \emph{Findings of the Association for Computational Linguistics: EMNLP 2024}, 1794-1809. https://doi.org/10.18653/v1/2024.findings-emnlp.97
\bibitem[Zhang(2026)]{ref21} Zhang, Z., Liu, Q., Liang, S., Li, N., Hu, Z., Gao, W., Li, R., Huang, Z., Rutkowski, L., Yu, B., \& Tao, D. (2026). Controllable contamination detection for reliable LLM evaluation with statistical guarantees. \emph{Proceedings of the 64th Annual Meeting of the Association for Computational Linguistics (Volume 1: Long Papers)}, 30122-30143.
\bibitem[Zhao(2024)]{ref22} Zhao, Q., Huang, Y., Lv, T., Cui, L., Sun, Q., Mao, S., Zhang, X., Xin, Y., Yin, Q., Li, S., \& Wei, F. (2024). MMLU-CF: A contamination-free multi-task language understanding benchmark. \emph{arXiv preprint} arXiv:2412.15194. https://doi.org/10.48550/arXiv.2412.15194
\bibitem[Zhu(2024)]{ref23} Zhu, W., Hao, H., He, Z., Song, Y.-Z., Yueyang, J., Zhang, Y., Hu, H., Wei, Y., Wang, R., \& Lu, H. (2024). CLEAN-EVAL: Clean evaluation on contaminated large language models. \emph{Findings of the Association for Computational Linguistics: NAACL 2024}, 835-847. https://doi.org/10.18653/v1/2024.findings-naacl.53
\end{thebibliography}
\end{document}